\documentclass[11pt,twocolumn]{article}
 
\usepackage[utf8]{inputenc}
\usepackage[T1]{fontenc}
\usepackage[margin=0.85in]{geometry}
\usepackage{amsmath,amssymb,amsthm}
\usepackage{mathtools}
\usepackage{bm}
\usepackage{xcolor}
 
\usepackage{booktabs}
\usepackage{multirow}
\usepackage{graphicx}
\usepackage{float}
\usepackage{caption}
\usepackage{subcaption}
\usepackage{placeins}
 
\usepackage[numbers,sort&compress]{natbib}
\usepackage[colorlinks=true,citecolor=blue,linkcolor=blue,urlcolor=blue]{hyperref}
 
\usepackage{enumitem}
 
\newcommand{\Field}{\mathcal{F}}
\newcommand{\Coh}{\widehat{\mathcal{R}}}
\newcommand{\Coheff}{\widehat{\mathcal{R}}^{\mathrm{eff}}}
\newcommand{\Cohext}{\widehat{\mathcal{R}}^{\mathrm{ext}}}
\newcommand{\dCoh}{\delta\widehat{\mathcal{R}}}
\newcommand{\Dscr}{\mathcal{D}}
\newcommand{\Kern}{\mathcal{K}}
\newcommand{\ConfigSpace}{\mathcal{C}(S,\lambda)}
\newcommand{\keff}{k_{\mathrm{eff}}}
\newcommand{\mueff}{\mu_{\mathrm{eff}}}
\newcommand{\cp}{\,\mathrm{cp}}
\DeclareMathOperator{\Var}{Var}
 
\theoremstyle{plain}
\newtheorem{axiom}{Axiom}
\newtheorem{postulate}{Postulate}

\newtheorem{uclaim}{Claim}

\theoremstyle{definition}

\begin{document}
 
\title{%
  \textbf{Information as Structural Alignment:}\\[4pt]
  \textbf{A Dynamical Theory of Continual Learning}%
}

\author{%
  \normalsize The Informational Buildup Foundation\\[4pt]
  \normalsize Radu Negulescu / April 2026\\[4pt]
  \normalsize radu@ibf.xyz  
}
\date{}
 
\maketitle
 
\begin{abstract}
\sloppy
Catastrophic forgetting is not an engineering failure. It is a mathematical consequence of storing knowledge as global parameter superposition. Existing methods (regularization, replay, frozen sub-networks) add external mechanisms to a shared-parameter substrate. None derives the retention mechanism from the learning dynamics themselves.

This paper introduces the Informational Buildup Framework (IBF), an alternative substrate for continual learning derived from the premise that information is the achievement of structural alignment rather than stored content. In IBF, two equations define the governing dynamics: a Law of Motion driving configuration toward higher coherence, and Modification Dynamics that persistently deform the coherence landscape in response to localized discrepancy signals. Memory, agency, and self-correction arise from these dynamics rather than being added as separate modules.

We demonstrate the full lifecycle on a two-dimensional toy model where every mechanism is visible, then validate across three domains: (1) a controlled non-stationary world, (2) chess (evaluated independently by Stockfish), and (3) Split-CIFAR-100 (20 tasks, frozen ViT encoder). 

Across all three domains, IBF achieves replay-superior retention without storing raw data. We observe near-zero forgetting on CIFAR-100 (BT = -0.004), positive backward transfer in chess (+38.5 cp), and 43\% less forgetting than replay in the controlled domain. In chess, the framework achieves a mean behavioral advantage of $+88.9 \pm 2.8\cp$ at the geometrically prescribed bandwidth under independent evaluation, exceeding MLP and replay baselines.

A key empirical result is regime dependence. The same agency mechanism yields three outcomes across discrepancy regimes: helpful in structured domains, harmful in contradictory contexts, and neutral when baselines are saturated, as predicted by the theory. Ablations confirm a developmental cascade: agency shapes which corrections are learned, activates the self-correction mechanism, and curated memory produces a systematic behavioral advantage.

\fussy
\end{abstract}

\vspace{1em}

\section{The Problem}
\label{sec:problem}

Train a neural network on task~A. It learns. Train it on task~B. It learns B, but performance on A collapses catastrophically. This is not a bug. It is a mathematical consequence of how the network stores knowledge. All weights are shared globally; learning B necessarily perturbs the parameter configuration that encoded A. Catastrophic forgetting~\cite{mccloskey1989catastrophic, french1999catastrophic} is structural, not incidental.

The response has taken the form of added mechanisms: regularization penalties~\cite{kirkpatrick2017overcoming}, frozen sub-networks, stored raw data~\cite{rolnick2019experience}, and growing task-specific capacity through methods such as dynamically expandable and progressive neural networks~\cite{yoon2018den,rusu2016progressive}. Each mitigates forgetting. None eliminates it by construction.

The cause runs deeper than any single method can reach. The dominant computational paradigm, whose early formalization traces to McCulloch and Pitts~\cite{mcculloch1943logical}, stores memory in shared parameters and updates knowledge through changes to those same parameters. There is no intrinsic mechanism ensuring that modifying one memory leaves others intact. These are not engineering oversights. They are consequences of the substrate itself.

The question, then, is this: is there a substrate where learning B need not destroy A by construction rather than by countermeasure?

This paper introduces one. It rests on a single premise: \emph{information is not data. It is the achievement of structural alignment between a system's internal configuration and the environment's structure.} From this premise, two equations define the governing dynamics. Memory, agency, and self-correction are then formulated as derived capacities of the framework: consequences the dynamics say should follow, rather than engineered additions layered on afterward.

Before stating the theory, we demonstrate it.

\section{The Toy Model: IBF in Two Dimensions}
\label{sec:toymodel}

We begin by instantiating the simplest possible learning system that exhibits every Informational Buildup Framework (IBF) mechanism \emph{operationalized in this paper}: a two-dimensional configuration space, two actions, and two sequential contexts. The system will learn from scratch, face a context switch that contradicts part of what it learned, and must sort the knowledge that transfers from the knowledge that does not. 

Every mechanism that appears later in the formal theory and the three-domain validation is visible here at a scale the reader can easily keep in mind.

\subsection{Setup}

Inputs are vectors $x \in \mathbb{R}^{2}$ sampled from a standard normal distribution. The system selects one of $k = 2$ actions for each input. The configuration space is $z = [x_1, x_2]$: a transparent encoding with no compression, so the reader sees exactly what the system sees. A baseline evaluator $\hat{\Coh}_{\mathrm{base}}(z)$ provides an initial assessment of each configuration, initialized near $0.5$ everywhere (the system starts knowing nothing).

The environment determines the correct action through a scoring function with two components:
\begin{equation}
  s_j(x, c) = \beta \cdot x_1 \cdot p_j
  + \alpha \cdot u_c \cdot x_2 \cdot r_j,
  \label{eq:toy_scoring}
\end{equation}
where $p = [+1, -1]$ and $r = [+1, -1]$ are coefficient vectors, $\beta = \alpha = 1.0$, and $u_A = +1$, $u_B = -1$.

This decomposition is the key to the entire demonstration. The first term ($\beta \cdot x_1 \cdot p_j$) is \emph{invariant}: it produces the same contribution in both contexts. When $x_1 > 0$, action~0 is favoured; when $x_1 < 0$, action~1 is favoured, regardless of context. The second term ($\alpha \cdot u_c \cdot x_2 \cdot r_j$) is \emph{context-specific}: it flips sign between Context~A and Context~B, creating maximal contradiction. A position dominated by $x_1$ has the same correct answer in both contexts. A position dominated by $x_2$ has opposite correct answers. The system does not know this decomposition. It must discover which aspects of the environment are universal and which are local from interaction alone.

\noindent\emph{Intuitively,} this toy world mixes truth and contradiction. One feature ($x_1$) expresses structure that remains valid across contexts, while the other ($x_2$) expresses structure that reverses when the context changes. The system is never told which is stable and which is local. It must infer that distinction from interaction alone, deciding what to preserve, what to isolate, and what to let dissolve.

\subsection{Step 1: The Empty Landscape}

Before any interaction, the system's internal coherence landscape is flat: it assigns nearly the same value to every location in
configuration space, with $\Coheff(z) \approx 0.5$ everywhere. The system has no basis for preferring one action over the other. It
selects at random. Performance is at chance ($\sim\!50\%$), as shown below.

\begin{center}
  \includegraphics[width=0.72\columnwidth]{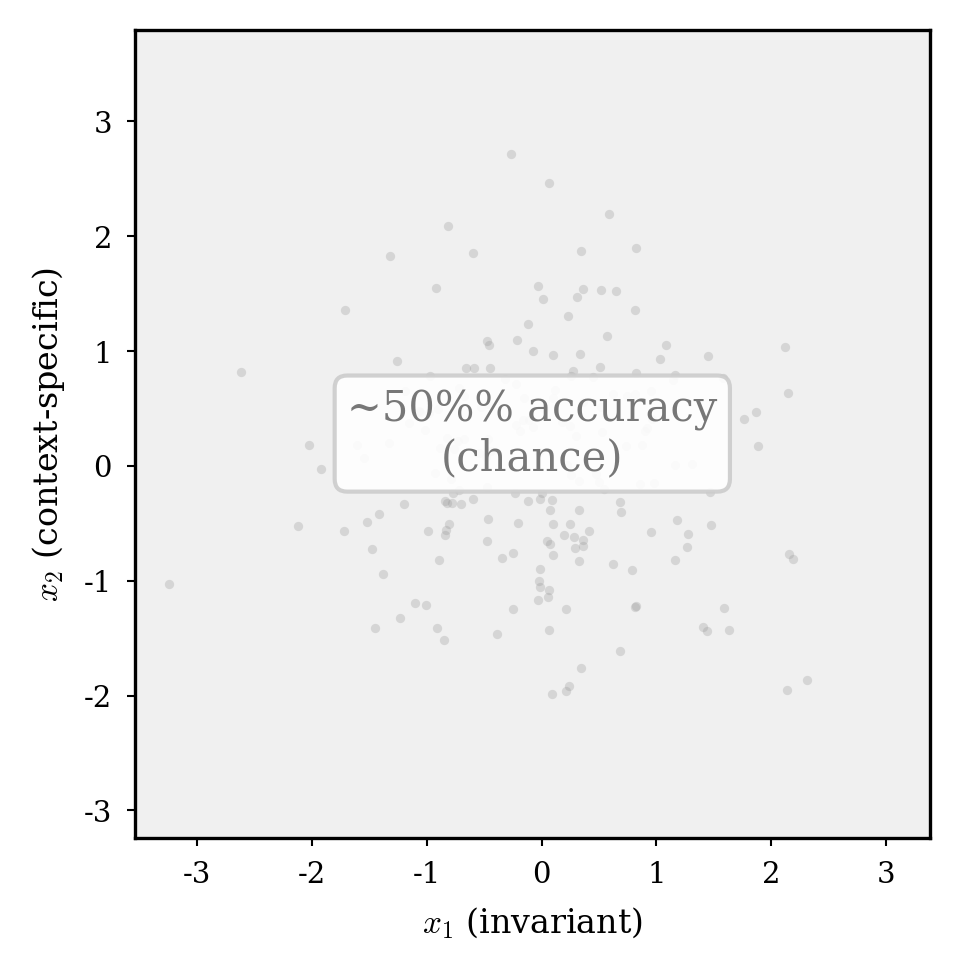}
  \captionsetup{font=footnotesize}
  \captionof{figure}{Before interaction: flat landscape, chance performance.}
  \label{fig:toy:a}
\end{center}

\subsection{Step 2: Interaction and Nucleation}

The system encounters its first inputs. For each input $x$, it selects action $a_j$ and receives a discrepancy signal
$\Dscr = R_{\mathrm{imposed}} - R_{\mathrm{chosen}}$, measuring the gap between the environmental truth and the system's current assessment. Where the gap is large and no existing memory center covers the visited region (kernel activation below threshold), a
new center \emph{nucleates} at $z$:

\begin{equation}
  c_{\mathrm{new}} = (z,\; v = \eta \cdot \Dscr,\; w = 0).
\end{equation}

The landscape deforms locally: a bump appears at $z$, shaped by the Gaussian kernel $\Kern(z, z_i) = \exp(-\|z - z_i\|^2 / 2\sigma^2)$. Nearby inputs now receive a corrected assessment. Distant inputs are unaffected. The system has begun to build a local map of where its evaluator is wrong, as shown below.

\begin{center}
  \includegraphics[width=0.72\columnwidth]{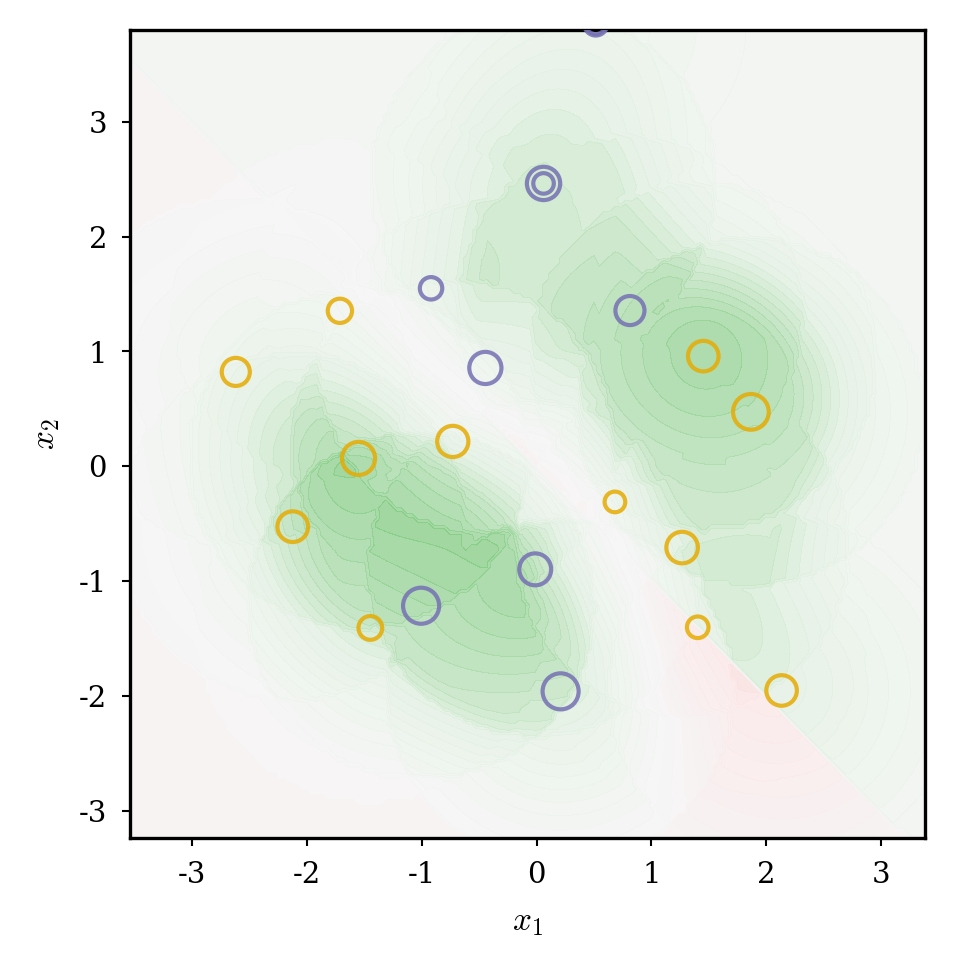}
  \captionsetup{font=footnotesize}
  \captionof{figure}{After 5 epochs: 22 memory centers nucleate at locations of large discrepancy signals. Green background indicates regions where the system chooses correctly; pink indicates errors.}
  \label{fig:toy:b}
\end{center}

\noindent\emph{Intuitively,} in IBF, learning begins where the system finds a mismatch between its current map and the world; where the map is already correct, there is nothing to correct.

\subsection{Step 3: Accumulation and Crystallization}

At the end of Phase~A (the first context), after 25 epochs ($5{,}000$ interactions), $23$ centers have nucleated. Each sits at a location the system visited repeatedly, encoding a correction to the baseline evaluator at that point.

Centers that receive repeated, consistent $\Dscr$-signals accumulate correction weight and eventually undergo a \emph{stability transition}: their decay rate drops by $60\times$ ($\mueff: 0.06 \to 0.001$). They stop fading. $18$ of $23$ centers crystallize. Some encode the invariant feature ($x_1$), others encode the context-specific feature ($x_2$). But the system does not yet know which is which. That distinction requires a second context.

Context~A accuracy rises from $0.50$ to $0.95$. The system has learned by deforming the coherence landscape, not by adjusting global parameters. Each correction is spatially localized: it improves performance near its location and has no effect elsewhere, as shown below.

\begin{center}
  \includegraphics[width=0.72\columnwidth]{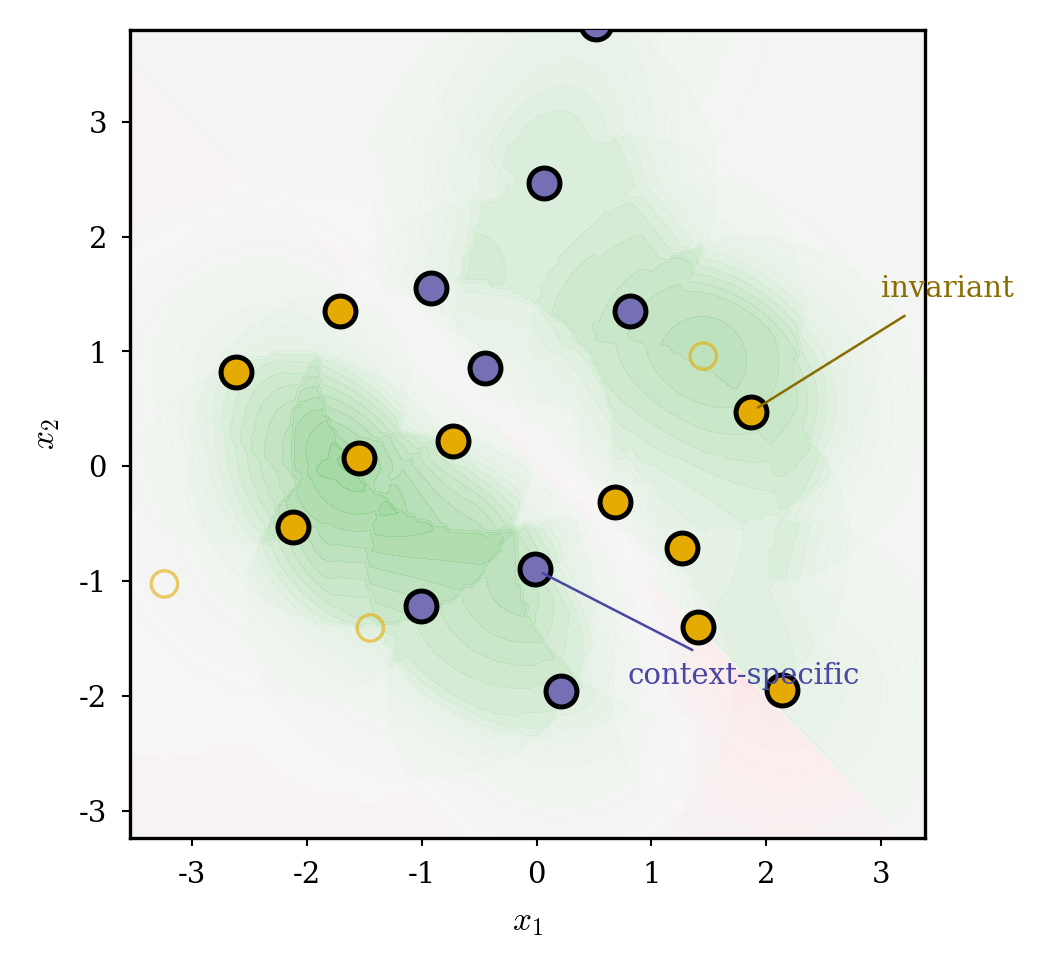}
  \captionsetup{font=footnotesize}
  \captionof{figure}{End of Phase~A: 18 of 23 centers crystallize (filled circles). Orange: invariant ($x_1$-encoding). Purple: context-specific ($x_2$-encoding).}
  \label{fig:toy:c}
\end{center}

\noindent\emph{Intuitively,} the system now carries stable memories from experience, but it does not yet know which ones are universal and which ones belong only to the current context. The next context will force that distinction.

\subsection{Step 4: Context Switch and Isolation}

Phase~B begins. The scoring function flips: $u_B = -u_A$. The $x_2$ component now demands the opposite answer from Phase~A. This is the moment where a standard neural network would catastrophically forget: training on Phase~B would overwrite the weight configuration that encoded Phase~A.

Here, something different happens. \emph{Context-gated read access} silences all Phase~A centers. They remain in the landscape but contribute nothing to the evaluation (the read-gating variable, $\gamma_i = 0$). They are asleep, not dead. New centers nucleate for Context~B, occupying their own regions of the space, as shown below.

\begin{center}
  \includegraphics[width=0.72\columnwidth]{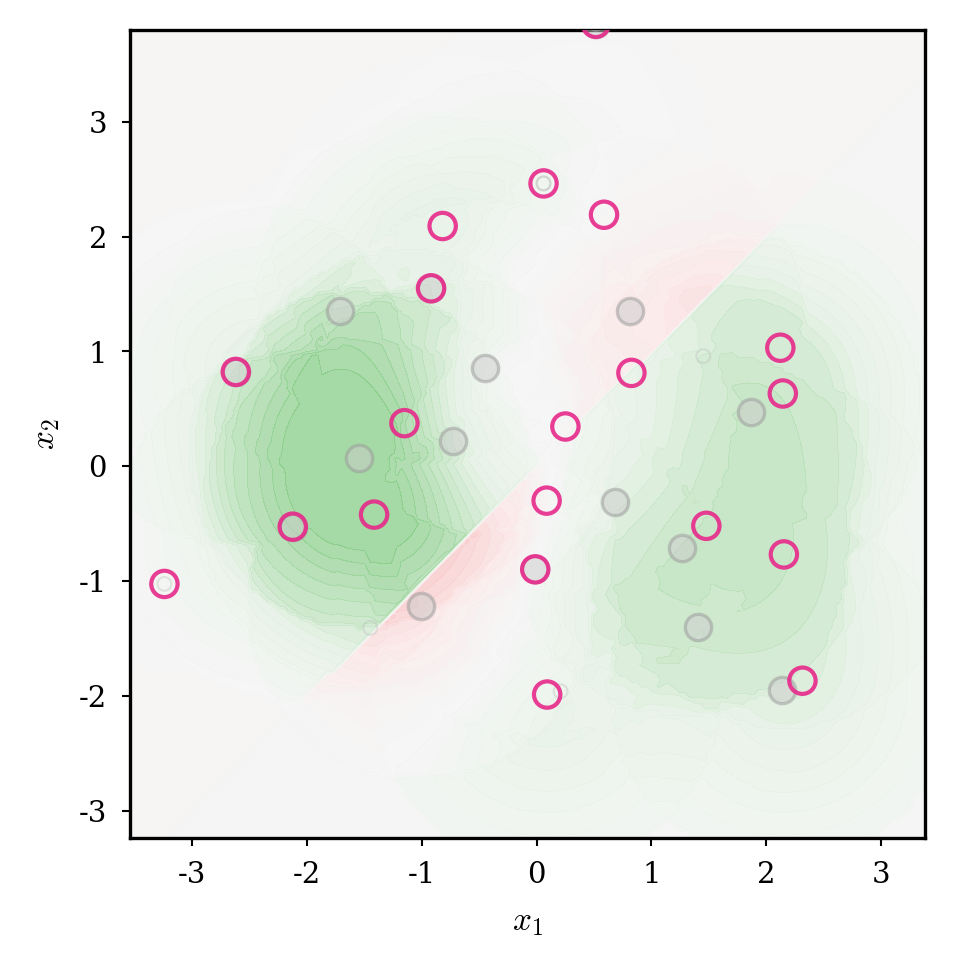}
  \captionsetup{font=footnotesize}
  \captionof{figure}{Previously crystallized centers (grey) remain in the landscape but are gated off, allowing new context-specific learning (pink) to begin without overwriting dormant memories.}
  \label{fig:toy:d}
\end{center}

\noindent\emph{Intuitively,} old knowledge is not destroyed when the world changes. It is temporarily silenced, creating space for new learning without erasing what came before.

\subsection{Step 5: The Crucible}

As Phase~B training proceeds, its inputs activate the Gaussian kernel near some Phase~A crystals. These sleeping crystals now face a test: does their stored correction agree with the new context's discrepancy signal?

\emph{Invariant crystals} ($x_1$-encoding): the raw $\Dscr$ from Phase~B is consistent with their stored correction. The $x_1$-structure they learned in Phase~A is still true in Phase~B. They survive. After accumulating at least $4$ cross-context updates without reversal, they earn \emph{broadcast rights}: $\gamma_i$ becomes $1$, projecting their corrections into Phase~B.

\emph{Context-specific crystals} ($x_2$-encoding): the raw $\Dscr$ from Phase~B contradicts their stored correction. The product $v_i \cdot \bar{\Dscr}_{\mathrm{raw}}$ falls below the reversal threshold ($-0.125$). They \emph{dissolve}: $\mueff$ reverts to $0.06$, broadcast rights are revoked, and they fade from the landscape.

Of $18$ Phase~A crystals, $14$ survive verification (predominantly invariant) and $4$ are dissolved (predominantly context-specific). The Crucible has separated universal from local knowledge using only raw cross-context discrepancy signals. No labels or task annotations were consulted. 

\begin{center}
  \includegraphics[width=0.72\columnwidth]{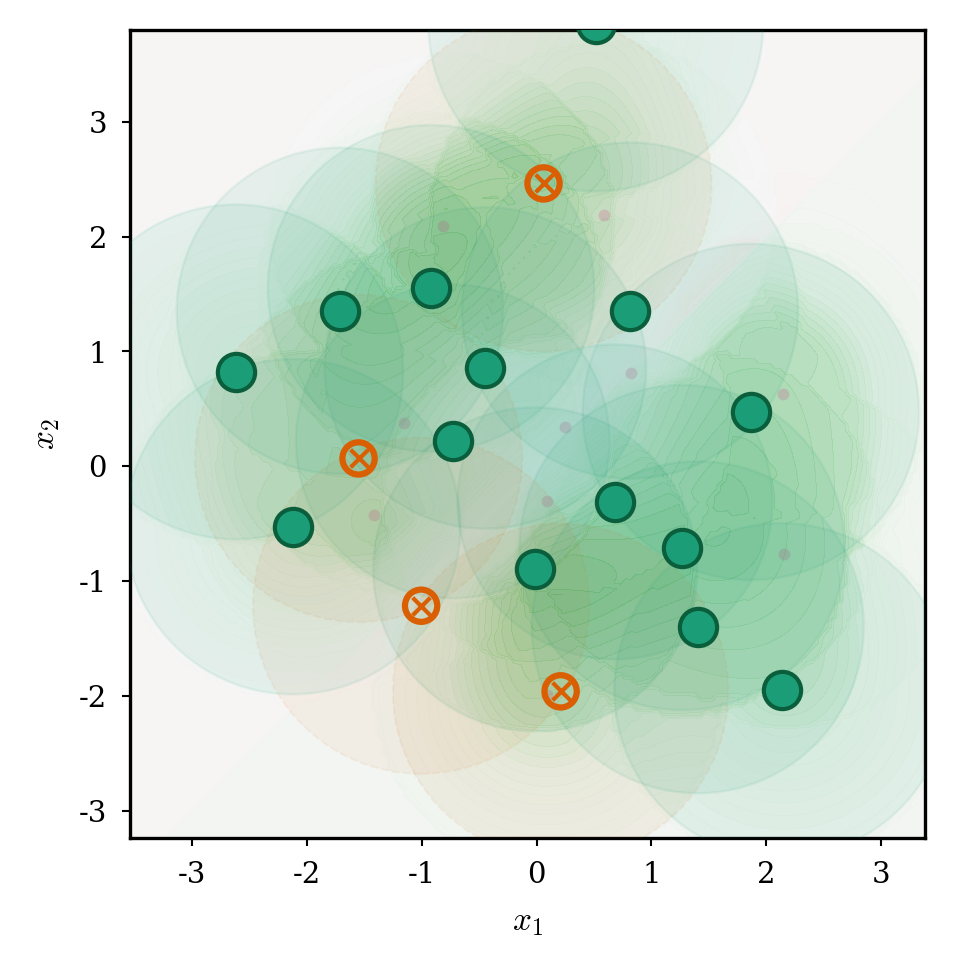}
  \captionsetup{font=footnotesize}
  \captionof{figure}{Universal corrections survive and earn broadcast rights (14, green), while context-specific corrections dissolve (4, orange).}
  \label{fig:toy:e}
\end{center}

\noindent\emph{Intuitively,} the system now discovers which memories were about the world itself and which were only about the old context. The first are kept and shared; the second are melted down.

\subsection{Step 6: Forward Transfer and Retention}

The $14$ verified universals now broadcast into Phase~B, contributing $x_1$-structure learned in Phase~A to a context that never trained on it directly. Phase~B reaches $0.97$ accuracy.

\begin{center}
  \includegraphics[width=0.72\columnwidth]{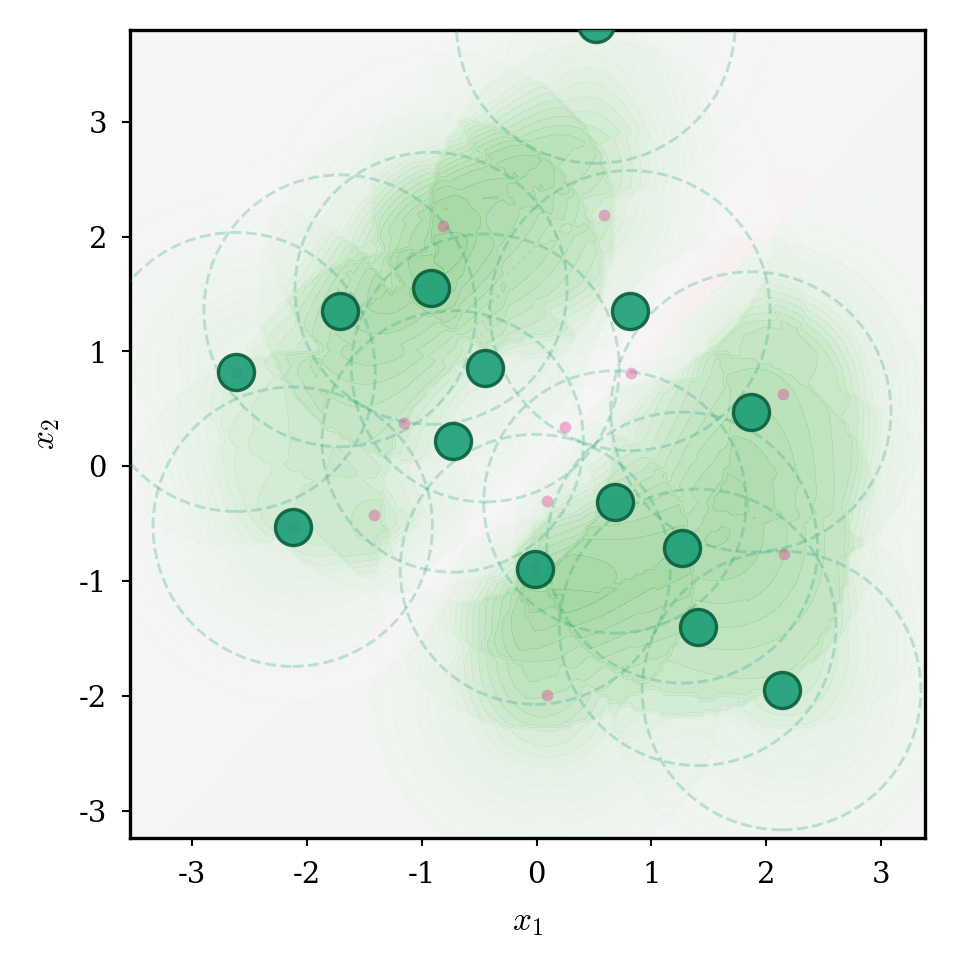}
  \captionsetup{font=footnotesize}
  \captionof{figure}{Verified universals broadcast into Phase~B (dashed circles: kernel influence).}
  \label{fig:toy:f}
\end{center}

\noindent\emph{Intuitively,} once a memory proves it reflects structure that remains true across contexts, the system can reuse it instead of learning that structure again from scratch.

\vspace{1em}

When we return to Phase~A inputs for evaluation, most crystallized corrections are intact. They survive because their decay rate dropped during crystallization ($\mueff = 0.001$), allowing them to physically resist the entropic passage of time and autonomously bridge hundreds of dormant epochs with near-zero erosion. 

But $\mathrm{Acc}_A = 0.778$ (vs.\ $0.952$ at end of Phase~A), giving $\mathrm{BT}_A = -0.174$. This is not zero forgetting. The Crucible's cross-context broadcast introduces some interference, the thermodynamic cost of forward transfer. The No-Crucible ablation confirms the tradeoff: without the Crucible, $\mathrm{BT}_A = 0.000$ (perfect isolation), but zero knowledge transfer between contexts.

\begin{center}
  \includegraphics[width=0.72\columnwidth]{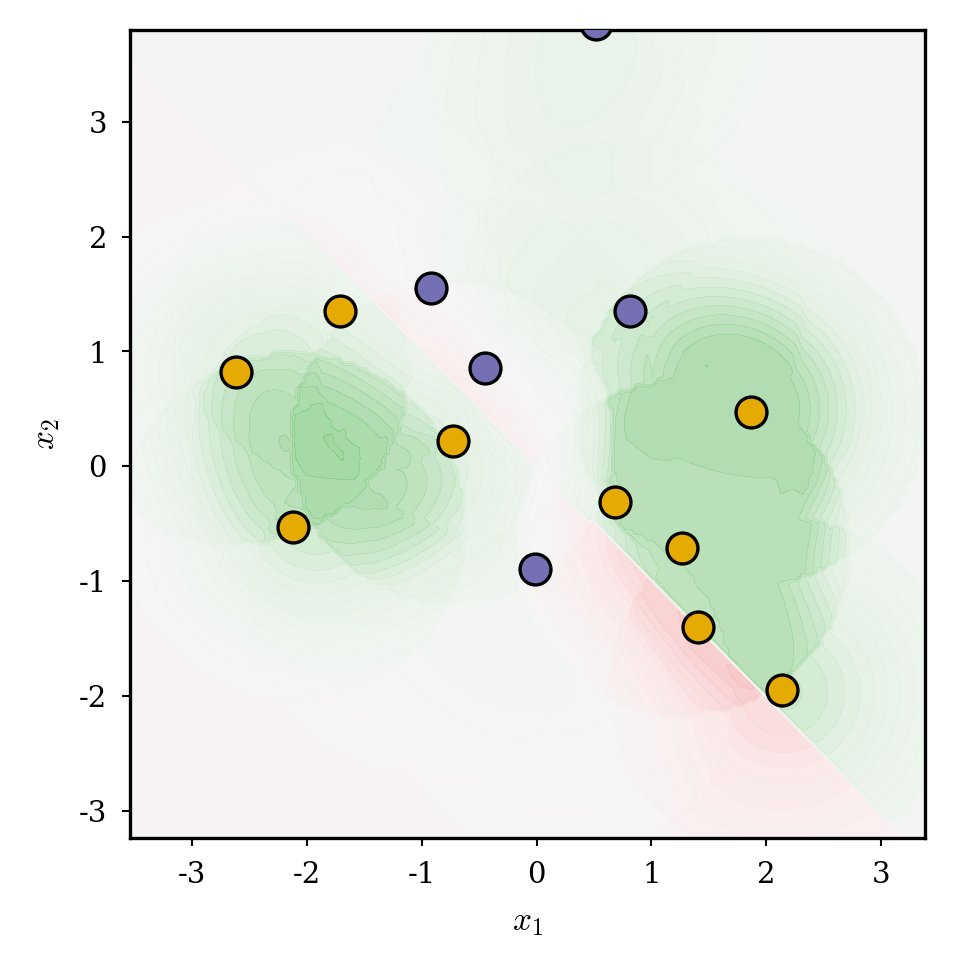}
  \captionsetup{font=footnotesize}
  \captionof{figure}{Return to Phase~A: $\mathrm{Acc}_A\!=\!0.778$ ($\mathrm{BT}_A\!=\!-0.174$). Previously learned corrections remain preserved in their original context, although cross-context broadcast introduces a limited transfer-for-retention tradeoff.}
  \label{fig:toy:g}
\end{center}

\noindent\emph{Intuitively,} in a world where contexts contradict each other, transfer comes with a cost: sharing knowledge helps the new context but introduces some interference with the old one. In domains where contexts share deeper structure, such as chess, the same mechanism can preserve prior knowledge more strongly and even improve it.

\subsection{Step 7: Emergent Agency}

Alongside the value corrections, a second modification channel has been accumulating throughout training. Each crystallized center accumulates an agency weight $w_i$ based on the variance of its $\Dscr$-signal history. Centers in low-variance regions, where corrections have been consistently reinforced, develop positive $w_i$, driving $\keff$ above baseline: the system commits to what it knows. Centers in high-variance regions, where corrections have been noisy or contradicted, keep $w_i$ low: the system hedges.

After Phase~B, $\keff$ varies from $5.0$ to $9.8$ across the $2$D space. The system has learned not just \emph{what} to correct but \emph{where to trust its own corrections}. This is the toy model's agency channel: spatially varying responsiveness that emerged from interaction rather than from a separate module.

\begin{center}
  \includegraphics[width=0.72\columnwidth]{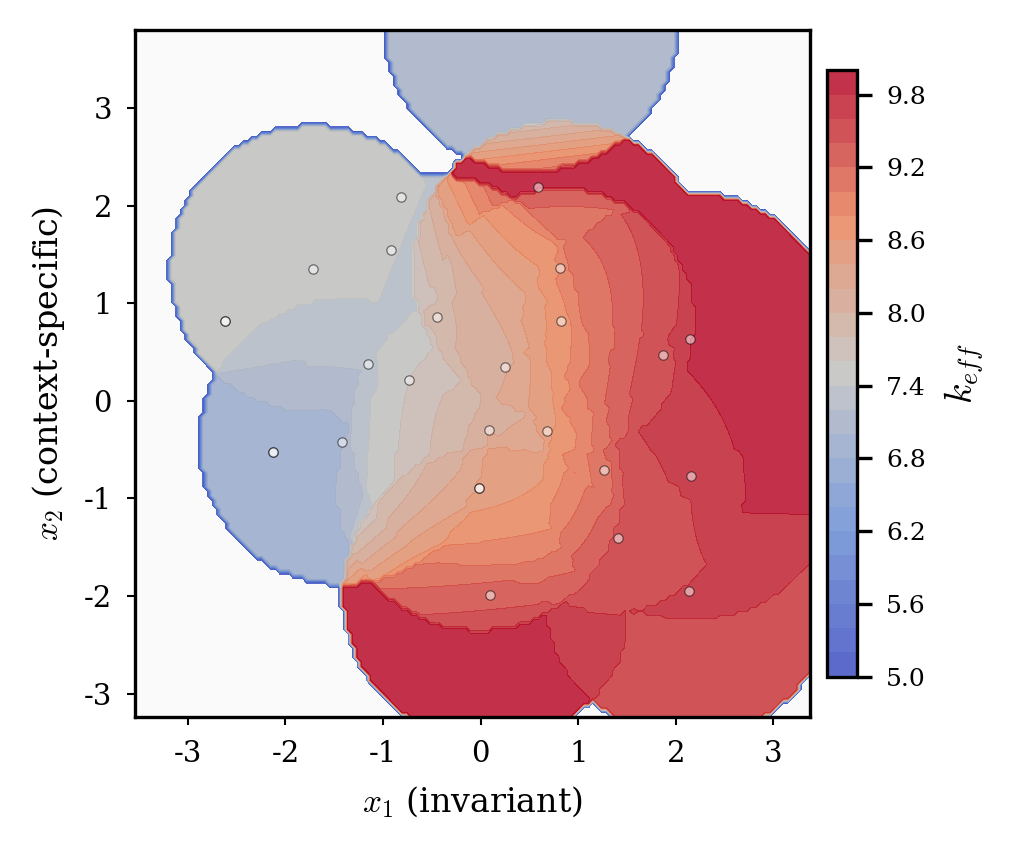}
  \captionsetup{font=footnotesize}
  \captionof{figure}{Emergent agency: $k_{\mathrm{eff}}$ ranges from $5.0$ to $9.8$, making responsiveness spatially nonuniform, increasing where corrections are reliable and remaining low where the local structure is uncertain or contradictory.}
  \label{fig:toy:h}
\end{center}

\noindent\emph{Intuitively,} experience is now shaping confidence itself. The system learns not only what tends to be true, but also where it should commit more strongly and where it should remain cautious.

\subsection{What You Just Watched}

Every mechanism in these seven steps (nucleation, crystallization, context gating, dissolution, broadcast verification, forward transfer, retention, and agency modulation) arose from two equations and one premise. No mechanism was added by hand. Each emerged as a consequence of the dynamics preserving structural alignment under non-stationary interaction.

\paragraph{Even at $d = 2$, the mechanisms differentiate.} A mini-ablation confirms that each mechanism produces a distinct behavioral signature in this minimal system:
\begin{center}
\small
\begin{tabular}{@{}lccc@{}}
\toprule
\textbf{Condition} & $\mathrm{Acc}_A$ (after B) & $\mathrm{BT}_A$ & $\mathrm{Acc}_B$ \\
\midrule
Full IBF    & 0.778 & $-0.174$ & 0.970 \\
No-Agency   & 0.766 & $-0.188$ & 0.960 \\
No-Cryst.   & 0.836 & $-0.132$ & 0.960 \\
No-Crucible & 0.952 & $+0.000$ & 0.950 \\
Passive     & 0.502 & $+0.000$ & 0.488 \\
\bottomrule
\end{tabular}
\end{center}

\noindent The No-Crucible condition confirms that the substrate itself does not forget: $\mathrm{BT}_A = 0.000$. The backward transfer observed in the full system ($-0.174$) is the thermodynamic cost of enabling cross-context transfer through the Crucible. The system can always choose perfect isolation. The question is whether the transfer justifies the cost. In this toy world, where contexts maximally contradict, the cost is visible. In domains where contexts share deeper structure, the same mechanism produces positive backward transfer: training on new contexts strengthens, rather than degrades, prior knowledge. The three-domain validation will show both regimes.

The toy model is not merely an illustration of the theory. It is the theory instantiated at $d = 2$. What follows in Section~\ref{sec:theory} gives it formal names. What follows in Sections~\ref{sec:engine}--\ref{sec:results} shows it working in three domains at $8$, $12$, and $68$ dimensions.

But first, we name what the reader has already seen.

\section{The General Theory}
\label{sec:theory}

The mechanisms observed in Section~\ref{sec:toymodel} are not engineered heuristics. They are consequences of one premise and two coupled equations: information as structural alignment, a Law of Motion, and Modification Dynamics. 

This section states that premise, the axioms of existence and motion, the modification postulate, and the derived claims that predict each observed capacity.

The full framework extends beyond the subset presented here, comprising a broader axiomatic, postulatory, and proof-level structure to be developed in a separate foundational work. For reference, two compact tables at the end of the paper summarize the formal primitives (Table~\ref{tab:primitives}) and situate the full theoretical arc, including the elements not stated in detail here (Table~\ref{tab:formal_summary}). For the purposes of the continual learning experimental program, however, we present only the essential components required in this paper. 

We begin with the fundamental premise from which the rest of the framework unfolds.

\subsection{The Alignment Premise}
\label{sec:theory:premise}
 
\begin{quote}
\emph{Information is not data. It is the achievement of structural alignment between a system's internal configuration and the structure of its environment.}
\end{quote}
 
Under this premise, memory becomes not storage, but persistent geometric deformation. Learning becomes the process of deformation. Forgetting, the erosion of deformation under entropic decay. And Intelligence becomes the capacity to deform systematically in ways that improve future alignment. 

From this premise, four axioms define the dynamics of existence and motion. One postulate defines the dynamics of modification. And the mechanisms observed in the toy model are consequences of this structure.

\subsection{Axioms of Existence and Motion}
\label{sec:theory:axioms}
 
\begin{axiom}[The Field]
\label{ax:field}

There exists a universal informational domain $\Field$ within which all patterns arise, persist, and dissolve. For any system $S$ at observational scale $\lambda$, the Field admits an associated differentiable configuration space $\ConfigSpace$ comprising all admissible internal states of the system.
\end{axiom}
\noindent\emph{Seen in toy model as} the 2D space $z = [x_1, x_2]$ (Figure~\ref{fig:toy:a}).

\vspace{0.5em}

\begin{axiom}[Coherence]
\label{ax:coherence}
For every system $S$ at observational scale $\lambda$, there exists a continuously differentiable function $\Coh(\cdot, S, \lambda): \ConfigSpace \to \mathbb{R}^{+}$ that quantifies the degree of structural alignment between the system's internal configuration and its environment. Coherence is relational (defined relative to a system and observational scale), fragile (generic unstructured perturbations degrade it), and non-decomposable (not globally expressible as a function of independent subsystem coherences).
\end{axiom}

\noindent\emph{Seen in toy model as} the baseline evaluator $\hat{\Coh}_{\mathrm{base}}(z)\approx 0.5$. As corrections accumulate, coherence basins form around reliable corrections (Figures~\ref{fig:toy:b}--\ref{fig:toy:c}).

\vspace{0.5em}

\noindent\emph{Intuitively,} coherence assigns a score to every point in the system's configuration space. That score measures how well the system's current internal configuration fits the structure of its environment. Higher coherence means better alignment; lower coherence means worse alignment. A stable system occupies a connected region where this fit remains high enough to persist.

\vspace{0.5em}

\begin{axiom}[Informational Gravity]
\label{ax:gravity}
Wherever coherence has a nonzero gradient, the system experiences a directional pull toward higher structural alignment:
\begin{equation}
  G(x_S) = \nabla_{x_S} \Coh(x_S, S, \lambda).
  \label{eq:gravity}
\end{equation}
\end{axiom}

\noindent\emph{Seen in toy model as} the discrete coherence increments $\{\Delta R^{\mathrm{eff}}_j\}$ across the two actions (Step~2).

\vspace{0.5em}

\noindent\emph{Intuitively,} coherence defines a landscape of better and worse fit. Informational gravity is the local pull of that landscape: at any point, it indicates the direction in which alignment improves most strongly.

\vspace{0.5em}

\begin{axiom}[Law of Motion]
\label{ax:motion}
The system evolves through configuration space along the gradient of its effective coherence landscape, modulated by a positive responsiveness field $k_S(x_S,S,\lambda) > 0$:
\begin{equation}
  \frac{dx_S}{dt} = k_S(x_S, S, \lambda)\,
  \nabla_{x_S}\Coheff(x_S, S, \lambda),
  \label{eq:law_of_motion}
\end{equation}
where $\Coheff = \Coh + \dCoh$ is the experience-modified coherence landscape. On a fixed landscape, the system ascends monotonically; apparent decreases in coherence occur only when the landscape itself is modified beneath the system's trajectory.
\end{axiom}

\noindent\emph{Seen in toy model as} Boltzmann action selection (Eq.~\ref{eq:softmax}). Responsiveness $\keff(z)$ controls the degree of determinism (Figure~\ref{fig:toy:h}).

\vspace{0.5em}

\noindent\emph{Intuitively,} the system does not merely know where better alignment lies; it moves in that direction. The coherence gradient tells it which way improves fit, while the responsiveness field $k_S$ determines how sharply or cautiously the system follows that direction.

\subsection{The Modification Postulate}
\label{sec:theory:postulate}

Axioms~\ref{ax:field}--\ref{ax:motion} describe motion on a fixed landscape. Learning requires the system to deform that landscape through interaction with signals external to its own gradient dynamics.

\begin{postulate}[Modification Dynamics]
\label{post:modification}
The effective coherence landscape evolves under localized discrepancy signals:
\begin{equation}
  \frac{\partial}{\partial t}\dCoh(y) =
  \eta\,\Kern(y, x_S)\,\Dscr(x_S) - \mu\,\dCoh(y),
  \label{eq:modification}
\end{equation}
where $\Dscr = \Cohext - \Coheff$ measures the discrepancy between the coherence structure imposed by the environment and the system's current effective evaluation, $\Kern$ is a localization kernel concentrated near the visited configuration $x_S$, $\eta$ governs the modification rate, and $\mu$ governs decay. A parallel equation governs the responsiveness modification $\delta k_S$ (the agency channel).

This postulate is subject to four structural constraints:
\emph{(i)}~the modification rate is slow relative to motion ($\eta \ll k_S$; timescale separation);
\emph{(ii)}~total modification remains bounded (capacity constraint);
\emph{(iii)}~the driving signal $\Dscr$ is structurally independent of the gradient that governs motion (separation from motion dynamics);
\emph{(iv)}~nontrivial interaction generically produces nonzero responsiveness modification (generic responsiveness modification).

An environment supports coherent modification only if its discrepancy signals exhibit nontrivial variation:
\begin{equation}
  \Var\bigl[\Dscr(x_S(t))\bigr] > 0
  \;\;\text{for a positive-measure set of}\; t.
  \label{eq:stable_structure}
\end{equation}
\end{postulate}

\noindent\emph{Seen in toy model as} Steps~2--3. Each interaction deposits a local correction shaped by $\Kern$. Corrections accumulate, building the landscape; they fade if unreinforced (transient centers decay at rate $\mu$), and crystallize when the local $\Dscr$-signal converges ($\mu \to 0$, producing long-lived structure).

\vspace{0.5em}

\noindent\emph{Intuitively,} motion alone lets the system climb a landscape, but learning requires the landscape itself to change. Modification Dynamics are the rule by which repeated mismatch locally reshapes that landscape, turning experience into memory.

\vspace{0.5em}

\noindent\textbf{Crystallization and dissolution.}
The decay rate $\mu$ is not fixed. When the local $\Dscr$-signal converges under consistent reinforcement, $\mu \to 0$ and the modification becomes long-lived structure (crystallization). When cross-context $\Dscr$-exposure systematically contradicts a crystallized modification, $\mu$ returns to its baseline value and the modification decays (dissolution). Both transitions are consequences of the modification dynamics acting on the local thermodynamic state of individual modification sites.

\vspace{0.5em}

\noindent\emph{Seen in toy model as} Step~5. Of 18 crystallized corrections, 4 are contradicted by Phase~B evidence and dissolve. The remaining 14, whose corrections are consistent with the new context, survive and earn broadcast rights.

\vspace{0.5em}

\noindent\emph{Intuitively,} memory stabilizes when the same correction keeps proving valid, and it loses that stability when later experience consistently contradicts it.

\subsection{Derived Capacities}
\label{sec:theory:claims}

From four axioms and one postulate, capacities conventionally engineered as separate modules appear here as derived consequences of the framework. The claims below state those capacities, ground them in the toy model, and identify their empirical signatures: the observable outcomes that would support or falsify them across the three validation domains. 

\vspace{0.5em}

\noindent\emph{These claims name, in formal terms, the capacities the reader has already seen emerge in the toy model.}

\vspace{0.5em}

\begin{uclaim}[Memory]
\phantomsection\label{thm:memory}
If a visited region reaches local thermodynamic equilibrium under Postulate~\ref{post:modification} (i.e., local $\Dscr$ converges), the induced modification undergoes a stability transition ($\mu \to 0$) and persists as a long-lived structure in the effective coherence landscape. Memory is preserved alignment expressed as persistent landscape deformation.
\end{uclaim}

\noindent\emph{Seen in the toy model as} crystallization (Step~3) and retention through context switch (Step~6).\\

\noindent\emph{Its empirical signature} is crystallized corrections surviving dormant epochs without replay, producing buffer-free retention across all three validation domains and, where applicable, backward transfer competitive with or superior to buffered baselines (\hyperref[sec:res:rrw]{\S\ref*{sec:res:rrw}}, \hyperref[sec:res:chess]{\S\ref*{sec:res:chess}}, \hyperref[sec:res:cifar]{\S\ref*{sec:res:cifar}}).\\

\noindent\emph{What would prove otherwise?} If crystallized modifications decay during dormant epochs despite the stability transition, or if they persist structurally but yield no measurable behavioral retention (with backward transfer indistinguishable from an unmodified baseline), then the Memory claim is false within this experimental program.

\vspace{0.5em}

\begin{uclaim}[Agency]
\phantomsection\label{thm:agency}
Under the responsiveness modification dynamics, nontrivial interaction generically produces spatially nonuniform $k_S(z)$: high where local discrepancy signals are consistent (low $\Dscr$-variance), and low where alignment is uncertain or contradictory (high $\Dscr$-variance). Agency is the interaction-acquired modulation of how strongly coherence gradients are followed. Under probabilistic action selection or concurrent landscape modification, this modulation produces differentiated trajectories; under deterministic evaluation on a fixed landscape, it does not.
\end{uclaim}

\noindent\emph{Seen in the toy model as} the $\keff$ heatmap (Figure~\ref{fig:toy:h}).\\

\noindent\emph{Its empirical signature} is the emergence of spatial structure in $\keff$ correlated with positional complexity ($\rho_{\mathrm{partial}}(\keff, H_{\mathrm{norm}}) > 0$). Removing agency degrades performance in structured-discrepancy domains (chess) while leaving saturated domains effectively unchanged (CIFAR), thereby producing regime-dependent outcomes from a single mechanism (\hyperref[sec:res:chess]{\S\ref*{sec:res:chess}}, \hyperref[sec:disc:demonstrated]{\S\ref*{sec:disc:demonstrated}}).\\

\noindent\emph{What would prove otherwise?} If heterogeneous discrepancy histories do not induce spatially nonuniform $k_S$, or if removing agency leaves exploration, cross-context exposure, and downstream performance unchanged in structured environments, then the Agency claim is false within this experimental program.

\vspace{0.5em}

\begin{uclaim}[Intelligence]
\phantomsection\label{thm:intelligence}
When memory and agency coexist, a system navigating an experience-refined landscape with agency-modulated responsiveness achieves systematically higher-alignment outcomes than the unmodified baseline, provided the environment contains nontrivial coherence structure (Eq.~\ref{eq:stable_structure}).
\end{uclaim}

\noindent\emph{Seen in the toy model as} accuracy increasing from $0.50$ to $0.95$ (Phase~A).\\

\noindent\emph{Its empirical signature} is that the fully adapted system achieves a statistically significant behavioral advantage over the passive baseline under independent evaluation (Stockfish in chess, held-out accuracy in CIFAR), and exceeds neural baselines where tested, including both MLP and replay in chess (\hyperref[sec:res:chess]{\S\ref*{sec:res:chess}}, \hyperref[sec:res:cifar]{\S\ref*{sec:res:cifar}}).\\

\noindent\emph{What would prove otherwise?} If, in environments satisfying Eq.~\ref{eq:stable_structure}, the joint presence of memory and agency does not yield systematically higher-alignment outcomes than the unmodified baseline under independent evaluation, then the Intelligence claim is false within this experimental program.

\vspace{0.5em}

\begin{uclaim}[Self-Correction]
\phantomsection\label{thm:reflexive}
A system with internally gated read access and cross-context verification can test the continued validity of its own memory. When later experience systematically contradicts a stored modification, that modification loses stability: its decay rate returns to the transient regime, broadcast rights are revoked, and its representational capacity returns to the Field. Self-correction is the system's ability to let false stability die.
\end{uclaim}

\noindent\emph{Seen in the toy model as} context gating and the Crucible (Steps~4--5). Of 18 Phase~A crystals, 14 survive verification and 4 dissolve.\\

\noindent\emph{Its empirical signature} is that cross-context exposure selectively destabilizes contradicted crystals while preserving verified ones. The dissolution rate is agency-dependent: without agency-driven exploration, the thermodynamic pressure that triggers selective dissolution does not develop (\hyperref[sec:res:chess]{\S\ref*{sec:res:chess}}, \hyperref[sec:disc:demonstrated]{\S\ref*{sec:disc:demonstrated}}).\\

\noindent\emph{What would prove otherwise?} If cross-context contradiction does not selectively destabilize contradicted modifications, leaving their dissolution, survival, and broadcast behavior indistinguishable from those of verified modifications, or if dissolution occurs indiscriminately in the absence of contradiction, then the Self-Correction claim is false within this experimental program.

\vspace{0.5em}
\begin{center}
\emph{---}
\end{center}

\noindent
Taken together, these four claims predict a developmental cascade. Persistent modification generates memory (\hyperref[thm:memory]{Memory}); interaction differentiates the system's responsiveness (\hyperref[thm:agency]{Agency}); cross-context verification enables the system to withdraw stability from memories that no longer preserve alignment (\hyperref[thm:reflexive]{Self-Correction}); and curated memory under agency-guided motion earns systematically better outcomes than the unmodified baseline (\hyperref[thm:intelligence]{Intelligence}).\\

\noindent
Removing any link degrades the capacities downstream of it. The ablation program in \hyperref[sec:experiments]{Section~\ref*{sec:experiments}} tests each link independently.

\subsection{The Discretization Bridge}
\label{sec:theory:discretization}

Any computational test of IBF operates in a finite regime: finite-dimensional latent spaces, discrete action sets, and finite modification sites. The following claim states the conditions under which that finite implementation converges to the continuous dynamics in the relevant respects.

\begin{uclaim}[Discrete Convergence]
\phantomsection\label{thm:discrete}
Under verifiable conditions on the encoder, coherence function, and implementation parameters, discrete IBF dynamics converge to Eqs.~\eqref{eq:law_of_motion} and~\eqref{eq:modification} in three respects:

\begin{enumerate}[label=(\alph*),leftmargin=2em,itemsep=1pt]
  \item \textbf{Representation (Condition~R):} The encoder produces a latent space in which $\Coh$ is $C^{2}$-smooth with sufficient absolute magnitude (Condition~R$'$).
  \item \textbf{Dynamics (Condition~A):} The discrete action set contains at least one action with positive coherence increment.
  \item \textbf{Modification:} The particle approximation of $\dCoh$ converges to the continuous field as the number of modification sites $M \to \infty$ with appropriately scaled kernel bandwidth $\sigma$.
\end{enumerate}
\end{uclaim}

\noindent\emph{Across computational instantiations of the framework, operational checks} include representation-ranking tests, magnitude checks (Condition~R$'$), increment checks (Condition~A), and kernel-resolution diagnostics based on merge rate.

\noindent\emph{Intuitively,} the theory is continuous, but any implementation is finite. The discretization bridge states when a finite latent representation, a finite action set, and a finite collection of memory particles are still sufficient to preserve the same effective dynamics.

\vspace{1em}
\noindent
Next, section~\ref{sec:engine} translates these continuous equations into the discrete computational particles
that implement them. Section~\ref{sec:domains} instantiates the engine across three domains and verifies Conditions~R, R$'$, and~A in each before any training begins. Section~\ref{sec:experiments} states the experimental predictions that follow from the claims. Section~\ref{sec:results} reports whether the data confirmed or refuted them.

\section{The Universal IBF Engine}
\label{sec:engine}

This section translates the continuous dynamics of Section~\ref{sec:theory} into a discrete computational engine. Continuous modification fields are realized as finite sets of kernel-localized particles in latent space.

In the implementations studied here, this particle approximation is instantiated through two homologous populations: one carrying coherence corrections, the other carrying responsiveness modulation. The precise class layout may vary by domain, but the mechanism does not: gated kernel readout, localized discrepancy-driven writing, convergence-triggered crystallization, contradiction-triggered dissolution, and capacity-controlled merging.

Every element introduced below is domain-agnostic. It operates on a $d$-dimensional latent space $z \in \mathbb{R}^{d}$ equipped with a frozen encoder $\pi$ and a baseline evaluator $\hat{\Coh}_{\mathrm{base}}$. No domain-specific architectures, training protocols, or task scenarios enter at this level. The memory system is not updated by backpropagation through its own state, no optimizer acts directly on its stored corrections, and no replay buffer is required.

The mapping to theory is direct: the Field becomes a finite latent configuration space; coherence becomes a baseline evaluator together with an experience-modified effective landscape; the Law of Motion becomes a discrete read-and-select procedure; the Modification Postulate becomes a local write rule; and the derived capacities of Section~\ref{sec:theory:claims} appear as lifecycle transitions within a finite particle system.

This engine should therefore be read as a computational instantiation of the continual-learning subset developed in this paper, not as an exhaustive realization of the full framework.

\subsection{The Particle Approximation}
\label{sec:engine:particle}

The universal engine stores its modifications in a finite particle representation. In the implementations used here, this appears as two homologous populations in latent space: a coherence-correction population carrying amplitudes $v_i$, and a responsiveness population carrying amplitudes $w_i$. Both populations are localized in the same latent geometry and share the same thermodynamic logic.

For notational economy, we write a generic particle as $c_i = (z_i, a_i, \sigma_i, \mu_{\mathrm{eff},i}, \mathrm{ctx}_i, \ldots)$, where $a_i$ is $v_i$ in the coherence channel and $w_i$ in the responsiveness channel. Across both populations, each particle is described by the following state:

\begin{center}
\footnotesize
\begin{tabular}{@{}llp{4.2cm}@{}}
\toprule
\textbf{Field} & \textbf{Type} & \textbf{Role} \\
\midrule
$z_i$ & $\mathbb{R}^{d}$ & Location in latent space \\
$a_i$ & bounded scalar & Correction amplitude ($v_i$ or $w_i$) \\
$\sigma_i$ & $\mathbb{R}^{+}$ & Local kernel bandwidth \\
$\mu_{\mathrm{eff},i}$ & $\mathbb{R}^{+}$ & Effective decay rate \\
$\mathrm{ctx}_i$ & int & Birth context \\
$\Dscr$-history & list & Local discrepancy history \\
verified & bool & Cross-context broadcast status \\
\bottomrule
\end{tabular}
\end{center}

A particle is \emph{transient} at birth ($\mu_{\mathrm{eff},i} = \mu_{\mathrm{base}}$). When its local discrepancy history converges, it undergoes a stability transition to the \emph{crystallized} regime ($\mu_{\mathrm{eff},i} = \mu_{\mathrm{cryst}} \ll \mu_{\mathrm{base}}$). When sustained contradiction later appears, the particle loses stability and returns to the transient regime.

\subsection{The Read Path}
\label{sec:engine:read}

\noindent The read path is the engine-level realization of Axioms~\ref{ax:coherence}--\ref{ax:motion}: stored modifications are reassembled into an effective coherence landscape, and action selection follows that landscape under locally modulated responsiveness.

\paragraph{Kernel readout.}
Readout begins by reconstructing the effective landscape from stored particles. The coherence-correction population contributes additively:
\begin{equation}
  \delta\hat{R}(z)
  =
  \sum_i \gamma_i\, v_i\, \Kern(z, z_i),
  \label{eq:particle_v}
\end{equation}
where $\Kern(z,z_i) = \exp(-\|z-z_i\|^2 / 2\sigma_i^2)$ is a Gaussian radial basis kernel with per-particle bandwidth $\sigma_i$, and $\gamma_i \in \{0,1\}$ is the read-gating variable.

The responsiveness population is read out intensively rather than additively:
\begin{equation}
  \delta k(z)
  =
  \frac{\sum_i \mathbb{I}_{\mathrm{cryst},i}\,\gamma_i\,w_i\,\Kern(z,z_i)}
       {\sum_i \mathbb{I}_{\mathrm{cryst},i}\,\gamma_i\,\Kern(z,z_i)},
  \label{eq:particle_k}
\end{equation}
where $\mathbb{I}_{\mathrm{cryst},i}$ is $1$ for crystallized responsiveness particles and $0$ otherwise. When the denominator is zero, $\delta k(z)$ is defined to be $0$. This intensive readout keeps responsiveness bounded and makes it a local modulation rather than an unbounded sum.

\paragraph{Reflexive read-gating ($\gamma_i$).}
Not every stored particle is allowed to contribute. A particle is readable only under the following conditions:
\begin{equation}
  \gamma_i =
  \begin{cases}
    1 & \text{same-context,} \\
    1 & \text{cross-context, crystallized and verified,} \\
    0 & \text{otherwise.}
  \end{cases}
  \label{eq:read_gating}
\end{equation}
Unverified particles are silent by default. This is the engine-level realization of the context-gated read access introduced in the toy model. In a more continuous formulation, cross-context readability would likely vary smoothly with a particle's thermodynamic state; here it is discretized deliberately as a minimal computational approximation.

\paragraph{Effective coherence and action selection.}
Once readable particles have been assembled into $\delta\hat{R}(z)$ and $\delta k(z)$, the engine evaluates candidate next states against the resulting effective landscape. The effective coherence is $\Coheff(z) = \hat{\Coh}_{\mathrm{base}}(z) + \delta\hat{R}(z)$, and the effective responsiveness is $\keff(z) = \max(k_{\min},\, k_0 + \delta k(z))$. Let $z_{\mathrm{current}}$ denote the current state and let $z_j$ denote the candidate next state associated with action $a_j$. Define $s_j^{\mathrm{eff}} = \Coheff(z_j)$ as the effective coherence assigned to that candidate. The Law of Motion (Axiom~\ref{ax:motion}) is instantiated as Boltzmann selection over candidate next states:
\begin{equation}
  P(a_j \mid z_{\mathrm{current}})
  =
  \frac{\exp\!\bigl(\keff(z_{\mathrm{current}})\, s_j^{\mathrm{eff}}\bigr)}
       {\sum_m \exp\!\bigl(\keff(z_{\mathrm{current}})\, s_m^{\mathrm{eff}}\bigr)},
  \label{eq:softmax}
\end{equation}
where $\Delta R_j^{\mathrm{eff}} = s_j^{\mathrm{eff}} - \Coheff(z_{\mathrm{current}})$. When $\Coheff(z_{\mathrm{current}})$ is shared across all candidates, this is equivalent up to a common offset to Boltzmann selection over coherence increments $\Delta R_j^{\mathrm{eff}}$.

\vspace{1em}
\noindent\emph{Intuitively,} the read path answers two questions at once: what past corrections apply here, and how strongly should the system trust them? The first reshapes the landscape itself; the second governs how sharply the system moves across it.

\subsection{The Write Path}
\label{sec:engine:write}

\noindent The write path is the direct computational realization of Postulate~\ref{post:modification}. It specifies how localized discrepancy deforms the effective landscape, and how the same interaction stream also shapes the responsiveness channel.

\paragraph{The discrepancy signal.}
After each interaction step, the engine receives a discrepancy signal $\Dscr$ that measures the gap between the environment's imposed coherence structure and the system's current effective evaluation. Its concrete form varies by domain and is specified in Section~\ref{sec:domains}. At the engine level, only its structural role matters: it is the source of modification, but it is not the gradient that governs motion.

\paragraph{Two-pass update.}
Once discrepancy is available, writing proceeds in two logically distinct passes. The separation is necessary. The engine must distinguish between \emph{learning from the active context} and \emph{testing whether past corrections remain valid under a changed one}.

\vspace{1em}

\noindent\emph{Pass~1: cross-context contradiction testing.}
Crystallized particles from other contexts whose kernel activation exceeds an exposure threshold receive the \emph{raw, unattenuated} discrepancy signal in their history logs. This pass serves only to assess continued validity under contradiction, not to write new local structure.

\noindent\emph{Pass~2: same-context spatial learning.}
Particles from the active context update through localized kernel-weighted writing:
\begin{equation}
  v_i \leftarrow
  \mathrm{clip}\!\bigl(
  v_i + \eta_i\,\Kern(z_i, z_{\mathrm{visited}})\,\Dscr,\,
  -v_{\max},\, v_{\max}
  \bigr),
  \label{eq:v_update}
\end{equation}
where $\eta_i = \eta_{\mathrm{base}}$ for transient particles and $\eta_i = \eta_{\mathrm{cryst}}$ for crystallized ones. These particles append \emph{kernel-local} discrepancy to their history, that is, the quantity $\Dscr \cdot \Kern$. If no same-context particle exceeds the creation threshold, a new particle is nucleated at $z_{\mathrm{visited}}$.

This separation is an architectural invariant: raw cross-context discrepancy is used for Crucible verdicts, while local kernel-weighted discrepancy is used for spatial learning.

\paragraph{Responsiveness update.}
The same interaction stream also updates the responsiveness population, but now the target is variance-sensitive rather than value-corrective. Each crystallized responsiveness particle is moved toward a variance-derived target
\begin{equation}
  w_{\mathrm{target}}
  =
  \mathrm{clip}\!\left(
  w_{\max}\Bigl(1 - \frac{\Dscr_{\mathrm{var}}}{\theta_w}\Bigr),
  -w_{\max},\, w_{\max}\right),
\end{equation}
where $\Dscr_{\mathrm{var}}$ is the rolling variance of the particle's local discrepancy history. It is computed after excluding an initial transient window and then using a recent rolling window. Low variance drives $w_i$ upward; high variance drives it downward.

\vspace{1em}
\noindent\emph{Intuitively,} the write path turns experience into local structural change. One part of the write path learns what to preserve; the other tests whether what was preserved remains true under changed circumstances.

\subsection{The Lifecycle}

\noindent The lifecycle operations realize, in discrete form, the derived capacities stated in Section~\ref{sec:theory:claims}: persistence through crystallization, selective loss of false stability through the Crucible, and bounded thermodynamic organization under decay and merge.

At each epoch boundary, the engine applies four operations in order.

\paragraph{1. Decay.}
All particles in both populations undergo passive thermodynamic fading:
$v_i \leftarrow (1-\mu_{\mathrm{eff},i})\,v_i$, \quad
$w_i \leftarrow (1-\mu_{\mathrm{eff},i})\,w_i$.
Dormant memories are not frozen. They survive because crystallization makes their decay rate extremely small. Retention is therefore near-zero erosion, not zero decay.

\paragraph{2. Crystallization.}
A transient particle undergoes the stability transition
$\mu_{\mathrm{eff},i} \to \mu_{\mathrm{cryst}}$ when two conditions are jointly satisfied:
\emph{(i)} sufficient exposure
($n_i \ge n_{\mathrm{cryst,min}}$), and
\emph{(ii)} convergence of recent discrepancy history
($|\bar{\Dscr}_{\mathrm{recent}}| < \theta_{\mathrm{conv}}$).
Convergence means the local correction has stabilized under continued reinforcement.

\paragraph{3. The Crucible (dissolution).}
A crystallized particle with sufficient cross-context exposure
($n_{\mathrm{cross}} \ge n_{\mathrm{cross,min}}$) is tested against recent raw cross-context discrepancy. If the product of its stored correction and the mean recent raw cross-context discrepancy falls below a reversal threshold,
\begin{equation}
  v_i \cdot \bar{\Dscr}_{\mathrm{raw,recent}}
  <
  \theta_{\mathrm{rev}},
  \label{eq:reversal}
\end{equation}
the particle loses stability: its decay rate returns to
$\mu_{\mathrm{base}}$, its cross-context broadcast rights are revoked, and it re-enters the transient regime. If not contradicted, it remains verified for cross-context readout. In the present instantiation, verification is phase-local: cross-context broadcast status is reset at context transitions before contradiction testing begins again.

Dissolution is distinct from decay. Decay is passive background erosion; dissolution is an explicit state transition triggered by contradiction.

\paragraph{4. Merge and capacity control.}
Within each population, particles from the same context that lie within a dynamic spatial threshold are merged. The more-updated particle absorbs the less-updated one: corrections, histories, and thermodynamic traces are consolidated, and the merged center retains the prescribed kernel geometry. If capacity is still exceeded after merging, crystallized particles are kept first, and only the remaining transient particles are filtered, with preference given to those that have accumulated the strongest update history.

\vspace{0.5em}

\noindent\emph{Intuitively,} the lifecycle gives the engine its thermodynamics: weak corrections fade, converged ones stabilize, contradicted ones lose that stability, and redundant ones collapse into more efficient local structure.

\subsection{Reference Tables}
\label{sec:engine:tables}

For ease of consultation, four compact reference tables are provided at the end of the paper. Table~\ref{tab:engine_mapping} maps the formal theory of Section~\ref{sec:theory} to the universal engine of Section~\ref{sec:engine}. Tables~\ref{tab:primitives}, \ref{tab:formal_summary}, and~\ref{tab:engine_dictionary_1} provide compact reference summaries of the formal primitives, the broader theoretical arc, and the core symbols used in the discrete implementation.

\vspace{0.5em}
\begin{center}
\emph{---}
\end{center}

\noindent The engine is now fully specified. Section~\ref{sec:domains} instantiates this universal mechanism across three validation domains spanning 8 to 68 dimensions, and verifies that the conditions required by the theory are satisfied in each before training begins.

\section{Validation Domains and Domain-Specific Instantiations}
\label{sec:domains}

This section instantiates the universal engine of Section~\ref{sec:engine} across three validation domains spanning $8$- to $68$-dimensional latent spaces. The mechanism itself is unchanged across domains. What varies is the frozen representation, the baseline evaluator, the discrepancy signal, and the legal action structure through which motion is discretized. Behavioral differences are therefore attributed not to changes in the mechanism, but to differences in latent geometry and environmental coherence structure.

These domains are not meant to exhaust the broader scope of IBF. They are validation instantiations of the present continual-learning formulation, chosen to isolate the mechanism under progressively less controlled conditions.

Each domain plays a distinct role. \textbf{RRW} is a controlled non-stationary environment with analytically known structure, used to test the mechanism under explicit cross-context contradiction. \textbf{Chess} is a combinatorially complex domain evaluated by an external oracle, used to test whether the same dynamics scale to genuine strategic structure. \textbf{CIFAR-100} is a high-dimensional continual-learning benchmark built on frozen visual representations, used to test whether the correction dynamics remain stable and non-destructive at larger scale.

Before introducing the domain-specific details, we first describe the geometric calibration principle shared by all three. This calibration determines the training bandwidth directly from latent-space geometry, without grid search, and provides the common scale-setting rule under which the universal engine is instantiated in each domain.

\subsection{Geometric Resolution Calibration}
\label{sec:domains:grp}

The kernel bandwidth $\sigma$ is the engine's central geometric parameter. Standard continual learning methods face analogous scale-setting problems and typically resolve them by domain-specific search over multiple training runs. IBF replaces search with measurement.

Across all domains, the same calibration strategy determines the training bandwidth $\sigma^{*}$ from the geometry of the frozen latent space. Two ingredients are required: the effective dimensionality $d_{\mathrm{eff}}$ (participation ratio of the PCA spectrum) and an empirical separation statistic between semantically distinct latent configurations. The precise statistic depends on the domain's action structure: RRW uses the passthrough latent geometry after action-separation calibration; chess uses legal-move sibling distances in the move-augmented space; CIFAR uses class-augmented sibling distances for value corrections together with a companion $64$D agency scale. The shared principle is a bleed criterion: $\sigma^{*}$ must be small enough that a correction deposited at one configuration does not destructively activate at a semantically distinct neighbor.

\paragraph{The empirical scaling ratio $\kappa$.}
The quantity $\kappa = \sigma^{*} / \sqrt{d_{\mathrm{eff}}}$ summarizes this safe training geometry as a single number per encoder:

\begin{center}
\small
\begin{tabular}{@{}llcccc@{}}
\toprule
\textbf{Domain} & \textbf{Encoder} & $d_z$ & $d_{\mathrm{eff}}$ & $\sigma^{*}$ & $\kappa$ \\
\midrule
RRW   & Passthrough   &  8 & 4.0  & 0.89 & 0.45 \\
Chess & 5-head CNN    & 12 & 4.9  & 1.27 & 0.57 \\
CIFAR & ViT-B/16+PCA  & 68 & 28.9 & 5.0  & 0.93 \\
\bottomrule
\end{tabular}
\end{center}

\noindent Within a given encoder family, $\kappa$ may serve as a transferable geometric ratio, allowing $\sigma^{*}$ for related tasks to be estimated without retraining or search.

\paragraph{Near-optimality of prescribed geometry.}
After training at $\sigma^{*}$, the system can be evaluated across a range of $\sigma_{\mathrm{eval}}$ values without retraining as a diagnostic of geometric calibration. Across all three domains, the geometrically prescribed bandwidth lies near the empirical optimum: RRW peaks at scale $1.1$, chess shows only a modest gain above the prescribed value, and CIFAR peaks at scale $0.9$.

No grid search was performed in any domain. The training bandwidth was derived geometrically, and post-training sweeps serve only as diagnostics of near-optimality.

\vspace{0.5em}

\noindent With the shared calibration rule established, we now instantiate the same engine in each validation domain. The domains differ in structure and scale, but each is evaluated under the same formal logic: a frozen representation instantiates the configuration space, a fixed baseline evaluator supplies the initial coherence landscape, a domain-specific discrepancy signal drives the modification dynamics, and a geometrically calibrated bandwidth determines the local support of those dynamics.

\subsection{Domain~I: Rotating Rules World (RRW)}
\label{sec:domains:rrw}

RRW is the controlled validation domain. Its role is to test the mechanism under explicit, analytically specified cross-context contradiction, where the underlying structure is fully known and every phase transition can be interpreted directly.

The environment operates in an $8$-dimensional latent space,
$z = [x_{4D};\, a_{\mathrm{emb},4D}]$, where $x$ is a continuous random input vector and $a_{\mathrm{emb}}$ is a scaled one-hot action embedding produced by a frozen passthrough encoder. The environment unfolds across three sequential phases, A, B, and C, all defined analytically. Its score decomposes into an invariant component, a shared phase-weighted component, and a contextual $2$D component. Phase~B is the exact reversal of Phase~A on that contextual subspace, while Phase~C introduces a new random orientation. The result is a controlled source of cross-context contradiction with fully known structure.

The baseline evaluator $\hat{\Coh}_{\mathrm{base}}$ is a frozen sigmoid evaluator with random affine initialization, increased only as needed to produce nontrivial spread in latent space. The $\Dscr$-signal is binary: $R_{\mathrm{imposed}} = 1$ if the chosen action is correct and $0$ otherwise. Geometric calibration yields $\sigma = 0.89$ ($d_{\mathrm{eff}} = 4.0$, $\kappa = 0.45$). Training proceeds for $3$ phases $\times$ $25$ epochs $\times$ $1{,}000$ points.

\paragraph{Baselines.}
All methods receive the same frozen representation,
$\Dscr$-signal, and sequential training trajectory:
(i)~Passive ($k \equiv k_0$, no modifications);
(ii)~pure MLP ($8 \to 64 \to 1$, SGD);
(iii)~replay MLP (same architecture, $50$-sample buffer).

\subsection{Domain~II: Chess}
\label{sec:domains:chess}

Chess is the complexity validation domain. Its role is to test whether the same dynamics scale from analytically controlled contradiction to genuine strategic structure, under evaluation by an external oracle rather than by a built-in analytical score.

The latent space is $12$-dimensional: an $8$-dimensional frozen board representation from a $5$-head CNN (win probability, material, phase, mobility, king safety) concatenated with a $4$-dimensional move-feature vector encoding from-square, to-square, piece type, and capture status. Positions are drawn from elite games and grouped into three sequential contexts by explicit heuristics over material imbalance, legal move count, and piece count: materially imbalanced positions (A), quiet/balanced positions with many pieces (B), and restricted-mobility positions (C).

The baseline evaluator $\hat{\Coh}_{\mathrm{base}}$ is the encoder's frozen win-probability head. During training, a separate Stockfish instance at depth~4 provides the environmental truth signal; the $\Dscr$-signal measures the discrepancy between the agent's internal coherence evaluation and this external oracle. Evaluation uses Stockfish depth~8 on disjoint position sets. Geometric calibration yields $\sigma^{*} = 1.2693$ ($d_{\mathrm{eff}} = 4.9$, $\kappa = 0.5747$), together with a separate $8$D agency scale for the responsiveness channel. 

\paragraph{Baselines.}
(i)~Passive;
(ii)~vanilla MLP ($12 \to 64 \to 64 \to 1$, ${\sim}5{,}000$ parameters, Adam);
(iii)~replay MLP ( $3{,}000$-sample buffer);
(iv)~value-only ablation ($\delta R$-only, $k \equiv k_0$).

\subsection{Domain~III: CIFAR-100}
\label{sec:domains:cifar}

CIFAR-100 is the scaling validation domain. Its role is to test whether the same correction dynamics remain stable and non-destructive in a higher-dimensional continual learning benchmark built on frozen visual representations.

A frozen ViT-B/16 encoder pretrained on ImageNet is projected via unsupervised PCA to a $64$-dimensional image space. For the value channel, each image vector is concatenated with a deterministic $4$-dimensional class-feature vector, yielding a $68$-dimensional class-augmented latent space. In this instantiation, value corrections operate in the $68$D augmented space, while agency remains in the underlying $64$D image space.

The baseline evaluator $\hat{\Coh}_{\mathrm{base}}$ is a frozen $100$-class coherence head trained on the $64$D frozen embeddings. Split-CIFAR-100 divides the $100$ classes into $20$ sequential tasks of $5$ classes each. Training is class-conditioned within each task: for every image, the engine processes the true class together with task-local wrong classes as candidate class-conditioned states, injects the corresponding baseline coherence values, and uses imposed targets to generate the discrepancy signal that drives correction. CIFAR is therefore instantiated here as a sequential class-conditioned correction domain on top of a frozen visual representation.

Geometric calibration yields $d_{\mathrm{eff}} = 28.9$ and a value-space bandwidth of $\sigma = 5.0$ in the reported run, with a separately measured $64$D agency scale used by the responsiveness channel. In the present paper, CIFAR serves as the high-dimensional scaling test: it asks whether the correction layer can preserve a strong frozen baseline across long task sequences without destructive interference.

\paragraph{Readout alignment.}
CIFAR also imposes one domain-specific readout requirement. Because the coherence head outputs softmax probabilities, direct linear readout creates a scale mismatch between the probability-based prior and the energy-like correction field: a correction of order $\pm 0.05$ against a probability gap of $0.19$ is too small to exert stable influence in linear space. Evaluating instead in log-space, $\log(\hat{\Coh}_{\mathrm{base}}) + \delta\hat{R}$, resolves this mismatch. In the reported run, the same learned corrections shift from harmful linear interference ($\mathrm{BT}_{\mathrm{lin}} = -0.085$) to near-zero forgetting under log-space readout ($\mathrm{BT}_{\mathrm{log}} = -0.004$). The training dynamics and discrepancy signal are unchanged; only the downstream classification readout differs. In this domain, faithful evaluation requires reading the correction field in a space consistent with the softmax-based baseline evaluator.

\paragraph{Baselines.}
(i)~Coherence head alone (frozen, no corrections);
(ii)~vanilla MLP;
(iii)~replay MLP ($5{,}000$-sample buffer);
(iv)~EWC.
All baselines operate on the same frozen encoder and evaluator.

\vspace{0.5em}
\begin{center}
\emph{---}
\end{center}

\paragraph{Frozen LLM extension.}
The same engine was also instantiated on top of a frozen Mistral-7B-v0.1 language model, extending the validation regime from frozen visual embeddings to frozen language-model outputs. In this setting, the baseline evaluator is given by the model's softmax probabilities over four multiple-choice answers, while value corrections operate in a $72$-dimensional proposition space formed by projecting all-mpnet-base-v2 sentence embeddings from $768$D to $64$D via PCA and appending a deterministic $8$-dimensional entity-feature vector. As in CIFAR, agency remains in the underlying $64$D space. The dataset consists of fictional entities and facts absent from pretraining, organized into three sequential phases: knowledge injection, cross-domain isolation, and counterfactual override. Training is contrastive within each phase, using the correct answer together with three wrong answers as candidate propositions, and geometric calibration follows the same sibling-distance logic used above.

This language-model instantiation serves a complementary role to CIFAR. Whereas CIFAR tests whether the correction layer remains stable in a high-dimensional regime where the frozen baseline is already strong, the frozen-LLM setting tests whether the higher-order mechanisms become necessary in a regime where the correction layer must do most of the work itself, including retention under continued learning and revision under explicit contradiction. \emph{The corresponding empirical results belong to a separate companion write-up and are not analyzed as part of the present paper}.

\vspace{0.5em}
\begin{center}
\emph{---}
\end{center}

\noindent
Finally, these domain-specific instantiations fix the empirical setting of the paper: in each case, a frozen representation defines the configuration space, a baseline evaluator provides the initial coherence landscape, and IBF contributes the modification dynamics that operate within it.

The experiments below therefore test IBF as a dynamical theory of continual learning on top of a supplied encoder/evaluator pair: specifically, a theory of how learned structure can be corrected, preserved, selectively revised, and adaptively deployed under sequential exposure without destructive overwrite. They do not test a full end-to-end account of representation formation from raw interaction, nor the broader multi-agent and open-world arc of the framework; those are left for later work.

With the domains, calibration procedure, and instantiations now specified, the remaining question is empirical: which predicted signatures must appear, and what would their absence imply for the framework?

\section{Experimental Program}
\label{sec:experiments}

The core claims of Section~\ref{sec:theory} each specify an empirical signature and a falsification condition. At a high level, those claims concern four derived capacities of the framework: memory, agency, self-correction, and intelligence. This section organizes their predicted signatures into a three-stage experimental program, states what each stage is designed to test, and defines what failure at each stage would imply for the framework.

\subsection{Progressive Validation Logic}
\label{sec:exp:logic}

The three stages are ordered by interpretive dependency, not by difficulty. Stage~I (RRW) validates mechanism interactions under analytically known structure, where every phase transition can be traced to its theoretical origin. Stage~II (Chess) tests whether the same dynamics produce emergent behavior in a domain where coherence structure must be discovered rather than specified. Stage~III (CIFAR-100) tests whether the correction dynamics remain stable in a higher-dimensional regime ($d_{\mathrm{eff}} = 28.9$) where the baseline is already strong.

Each stage is interpretable only because the preceding stage has confirmed the mechanisms on which it depends. If Stage~I fails, the framework is falsified at the mechanistic level within this experimental program. If later stages fail, the scope of the framework is bounded accordingly.

\subsection{Stage~I: Mechanism Confirmation (RRW)}
\label{sec:exp:stage1}

RRW is the controlled mechanistic test. Three targeted ablations, No-Crystallization, No-Crucible, and No-Agency, isolate the causal contribution of the core mechanisms under five independent seeds. Baselines and domain details are specified in \S\ref{sec:domains:rrw}.

\vspace{1em}
\noindent
The stage tests three predictions:
\begin{enumerate}[label=\textbf{P\arabic*.}, leftmargin=2.2em, itemsep=0.5em, topsep=0.35em]
  \item \textbf{Retention Without Replay.} IBF achieves backward transfer competitive with or superior to both MLP baselines without replaying stored past observations. This tests the empirical signature of the Memory claim: crystallized corrections survive dormant epochs and produce buffer-free retention.

  \item \textbf{Crucible-Mediated Transfer.} The No-Crucible ablation achieves near-perfect context isolation ($\mathrm{BT}_A \approx 0$), confirming that the gating architecture is safe by default. The Crucible's contribution is to permit cross-context transfer at a retention cost.

  \item \textbf{Forward Learning Without Overwrite.} IBF achieves final-phase accuracy exceeding both baselines. Spatial localization allows new learning to occupy distinct regions of $z$-space rather than overwriting dormant crystals.
\end{enumerate}

\subsection{Stage~II: Emergence Validation (Chess)}
\label{sec:exp:stage2}

Chess tests whether the mechanisms confirmed in Stage~I produce emergent behavior under genuine complexity. Each prediction depends on a mechanism already isolated in Stage~I. Baselines, oracle separation, and domain details are specified in \S\ref{sec:domains:chess}.

\paragraph{Measurement.}
Behavioral quality is measured by Stockfish~16 at depth~8 under deterministic single-threaded evaluation. Training uses a separate Stockfish instance at depth~4 on disjoint positions, preventing train-test leakage. All agents receive identical training signals.

All significance tests use $n = 1{,}002$ evaluation positions as the unit of analysis within each run. Centipawn advantages are computed per position, and Wilcoxon signed-rank tests assess whether paired differences are significantly nonzero. Because all positions are evaluated from a single trained model per condition, the observations within a run share the learned correction landscape. The resulting analysis therefore measures whether that learned landscape produces systematic behavioral advantage across a diverse position sample. The main chess behavioral and backward-transfer results are reported across three independent seeds sharing the same frozen encoder and geometric calibration, while the more detailed mechanism comparisons in the Results section are shown for a reference run.

\paragraph{Evaluation metric convention.}
All centipawn evaluations across all agents and baselines are symmetrically clipped at $\pm 1000\cp$ prior to statistical aggregation. Positions beyond this range correspond to decisive or forced-mate outcomes, where the centipawn scale loses its continuous positional meaning. The clip ensures that measured behavioral advantages reflect systematic shifts in positional alignment rather than sampling variance from terminal-state outliers.

\vspace{1em}
\noindent
The stage tests three predictions:
\begin{enumerate}[label=\textbf{P\arabic*.}, start=4, leftmargin=2.2em, itemsep=0.5em, topsep=0.35em]
  \item \textbf{Behavioral Advantage Under Independent Evaluation.} The fully adapted IBF agent achieves a statistically significant centipawn advantage over the passive baseline. The No-Agency \emph{training} ablation then isolates the developmental contribution: training without spatially varying responsiveness produces a weaker correction landscape. This jointly tests the empirical signatures of the Intelligence and Agency claims: memory provides the corrections, and agency shapes which corrections are learned.

  \item \textbf{Emergent Spatial Agency.} The agent autonomously develops spatially nonuniform $\keff(z)$ correlated with positional structure ($\rho_{\mathrm{partial}}(\keff, H_{\mathrm{norm}}) > 0$). This tests the empirical signature of the Agency claim: nontrivial interaction produces spatially differentiated responsiveness.

  \item \textbf{Transfer and Retention Under Crucible Verification.} The Crucible produces measurable cross-context transfer: verified corrections carry knowledge across context boundaries, manifesting as both forward transfer (earlier structure improving later-context performance) and positive backward transfer (later training improving earlier-context performance through compatible corrections). Buffer-free retention is competitive with or superior to the replay baseline. This jointly tests the empirical signatures of the Memory and Self-Correction claim: crystallized corrections persist through dormant epochs, and the Crucible selectively dissolves contradicted modifications while preserving verified ones.
\end{enumerate}

\subsection{Stage~III: Scalability Validation (CIFAR-100)}
\label{sec:exp:cifar}

Stage~III tests whether the dynamics scale to a standard benchmark at $68$ dimensions. The frozen ViT-B/16 encoder provides a strong baseline ($90.1\%$); the question is whether IBF's correction layer can preserve that baseline across $20$ task transitions while the compared baselines degrade it. Baselines and domain details are specified in \S\ref{sec:domains:cifar}.

\vspace{1em}
\noindent
The stage tests two predictions:
\begin{enumerate}[label=\textbf{P\arabic*.}, start=7, leftmargin=2.2em, itemsep=0.5em, topsep=0.35em]
  \item \textbf{Near-Zero Forgetting at Scale.} IBF achieves substantially lower backward-transfer degradation than all conventional baselines across $20$ sequential tasks, with $\sigma$ determined by geometric calibration rather than search. This tests whether the Memory claim remains effective at high dimensionality under prolonged sequential exposure.

  \item \textbf{Regime-Dependent Agency.} The agency channel produces negligible effect in CIFAR's saturated regime, in contrast to its essential role in chess and its harmful effect in RRW. The same mechanism thus produces three qualitatively different outcomes across three $\Dscr$-signal regimes, supporting the interpretation that agency is a derived property of the dynamics rather than an engineered feature.
\end{enumerate}

\subsection{Falsification Criteria}
\label{sec:exp:falsification}

Each claim in Section~\ref{sec:theory:claims} specifies its own falsification condition. Here we state how failure at each experimental stage would constrain the framework within this experimental program.

\emph{Stage~I failure} (any of P1--P3 not confirmed): the mechanisms do not operate as specified under known structure. The framework is falsified at the foundational level within this experimental program.

\emph{Stage~I success, Stage~II failure} (P1--P3 confirmed, any of P4--P6 not confirmed): the mechanisms are sound but do not produce emergent behavior in complex domains. The framework is falsified at the mechanistic level within this experimental program.

\emph{Stages~I--II success, Stage~III failure} (P1--P6 confirmed, one or both of P7--P8 not confirmed): the dynamics produce emergent behavior but do not fully scale to high-dimensional benchmark settings. The scope of the framework is thereby bounded accordingly.

\emph{All three stages succeed}: the dynamics are mechanistically sound (Stage~I), capable of producing emergent intelligent behavior under genuine complexity (Stage~II), and stable at benchmark scale under high-dimensional sequential exposure (Stage~III).

\section{Results}
\label{sec:results}

The experimental program states what each domain is expected to confirm or refute. We now turn from prediction to evidence. The results are reported in the same staged order: first the controlled mechanistic validation in RRW, then the emergence validation in chess, and finally the scalability validation in CIFAR-100. This ordering matters. Later stages are interpretable only because earlier stages establish the lower-level mechanisms on which they depend.

\subsection{Stage~I Results: Mechanism Confirmation in RRW}
\label{sec:res:rrw}

RRW answers the three Stage~I predictions under analytically known structure. Because all methods receive the same frozen representation, discrepancy signal, and sequential training trajectory, the comparison isolates learning mechanism rather than information access. The five-seed results are summarized below.

\begin{table}[!htbp]
\centering
\caption{\textbf{RRW results (5 seeds, mean $\pm$ std).} Full IBF achieves substantially stronger backward transfer and forward learning than the compared MLP baselines. The ablations do not function here as weaker versions of the full system, but as controlled probes of distinct mechanisms inside an explicitly contradictory domain.}
\label{tab:rrw_results}
\vspace{4pt}
\footnotesize
\resizebox{\columnwidth}{!}{%
\begin{tabular}{@{}lccc@{}}
\toprule
\textbf{Condition} & $\mathrm{Acc}_A$ & $\mathrm{BT}_A$ & $\mathrm{Acc}_C$ \\
\midrule
Full IBF           & $.684 \pm .051$ & $-.234 \pm .052$ & $.913 \pm .008$ \\
No-Agency          & $.721 \pm .047$ & $-.198 \pm .052$ & $.918 \pm .006$ \\
No-Crystallization & $.767 \pm .103$ & $-.152 \pm .108$ & $.920 \pm .008$ \\
No-Crucible        & $.911 \pm .007$ & $-.005 \pm .003$ & $.921 \pm .003$ \\
MLP                & $.442 \pm .059$ & $-.522 \pm .066$ & $.489 \pm .098$ \\
Replay MLP         & $.536 \pm .121$ & $-.410 \pm .113$ & $.816 \pm .029$ \\
Passive            & $.251 \pm .010$ & $+.000 \pm .000$ & $.248 \pm .009$ \\
\bottomrule
\end{tabular}%
}
\end{table}

\FloatBarrier

RRW is intentionally adversarial. Its phases are constructed to contain \emph{maximal explicit contradiction}, not deep reusable structure. In this regime, the ablations are not simply weaker versions of the full system, but variants that face less contradiction. No-Crucible shows this most clearly: with cross-context broadcast disabled, retention becomes nearly perfect, but transfer disappears. Full IBF pays a measurable cost because it is the only condition that attempts to carry structure across phases and discover whether it remains valid there.

\paragraph{P1. Retention Without Replay.}
Full IBF achieves $\mathrm{BT}_A = -0.234 \pm 0.052$, a $43\%$ reduction in forgetting relative to Replay MLP ($-0.410 \pm 0.113$) and a $55\%$ reduction relative to the pure MLP ($-0.522 \pm 0.066$), without replaying stored past observations. Retention is therefore carried by crystallized local corrections rather than by parameter protection or buffer rehearsal. In this domain, stronger retention in some ablations comes from stricter context isolation.

\paragraph{P2. Crucible-Mediated Transfer.}
The decisive RRW ablation is No-Crucible: $\mathrm{BT}_A = -0.005 \pm 0.003$, indicating near-perfect context isolation. This confirms that safety is enforced by the gating architecture itself. The Crucible's role is not defensive but transfer-enabling: it allows cross-context reuse at a retention cost. In RRW, that cost is visible because the phases are deliberately contradictory. The domain therefore shows the Crucible in its \emph{low-transfer regime}: it can still test and promote structure, but there is very little genuinely universal structure to promote.

\paragraph{P3. Forward Learning Without Overwrite.}
Full IBF reaches final-phase accuracy $\mathrm{Acc}_C = 0.913 \pm 0.008$, exceeding Replay MLP ($0.816 \pm 0.029$) and far exceeding the pure MLP ($0.489 \pm 0.098$). The system can learn new structure without erasing what it had already learned. Later-phase corrections occupy distinct regions of latent space rather than overwriting dormant structure. In RRW, slightly higher Phase~C scores in the simpler ablations reflect stricter isolation and lower transfer pressure. Full IBF retains strong forward learning while also carrying the additional burden of cross-context testing and selective reuse.

\vspace{1em}

The No-Agency ablation slightly improves RRW retention relative to the full system ($-0.198$ vs.\ $-0.234$). In RRW, agency operates in a deliberately adverse regime: exploration pushes more corrections into contradictory cross-context territory, where they encounter destabilizing evidence. The sign reversal relative to chess is therefore informative. It shows that agency is not uniformly beneficial; its value depends on whether the environment contains reusable structure or mainly contradiction.

\subsection{Stage~II Results: Emergence Validation in Chess}
\label{sec:res:chess}

Chess tests whether the mechanisms isolated in Stage~I produce emergent behavior under genuine strategic complexity. The experiments use a frozen 5-head CNN encoder, a 12D move-augmented latent space, geometrically calibrated training bandwidth $\sigma^{*} = 1.2693$, and deterministic Stockfish~16 evaluation at depth~8 on $n = 1{,}002$ disjoint positions per run. All centipawn values are symmetrically clipped at $\pm 1000\cp$, as specified in Section~6.3. The main behavioral and backward-transfer results are reported across three independent seeds; more detailed mechanism comparisons refer to the reference run.

\FloatBarrier
\begin{table}[!htbp]
\centering
\caption{\textbf{Chess reference run at the geometrically prescribed bandwidth under independent evaluation.} Mean centipawn values are reported after deterministic greedy evaluation on $1{,}002$ disjoint positions at $\sigma_{\mathrm{eval}}=\sigma^{*}=1.2693$. The No-Agency row quantifies the developmental contribution of agency at the prescribed geometric scale.}
\label{tab:chess_results}
\vspace{4pt}
\footnotesize
\setlength{\tabcolsep}{4pt}
\begin{tabular}{@{}lcc@{}}
\toprule
\textbf{Agent} & \textbf{Mean cp} & \textbf{vs.\ Passive} \\
\midrule
Full IBF ($\sigma^{*} = 1.2693$) & $-243.0$ & $+90.2$ \\
No-Agency ($\sigma^{*}$) & $-251.2$ & $+81.9$ \\
MLP & $-250.9$ & $+82.3$ \\
Replay & $-256.8$ & $+76.4$ \\
Passive & $-333.2$ & $0.0$ \\
\bottomrule
\end{tabular}
\end{table}

\FloatBarrier
\paragraph{Seed replication.}
Across three independent seeds, evaluated at the geometrically prescribed bandwidth $\sigma_{\mathrm{eval}}=\sigma^{*}=1.2693$, IBF achieves a mean behavioral advantage of $+88.9 \pm 2.8\cp$ over the passive baseline, together with positive backward transfer of $\mathrm{BT}_A = +35.4 \pm 2.9\cp$ and $\mathrm{BT}_B = +0.3 \pm 3.6\cp$. The more detailed mechanism comparisons below are presented for the reference run.

\paragraph{P4. Behavioral Advantage.}
Across three independent seeds at the geometrically prescribed bandwidth, IBF achieves a mean behavioral advantage of $+88.9 \pm 2.8\cp$ over the passive baseline. In the reference run at the same prescribed bandwidth, the fully adapted IBF agent achieves $+90.2\cp$ over the passive baseline ($p < 10^{-4}$). The developmental contribution of agency appears only in the training ablation. Training without spatially varying responsiveness reduces the advantage to $+81.9\cp$, a drop of $8.3\cp$ relative to the full system. This matches the distinction predicted in Section~\ref{sec:exp:stage2}: memory provides the corrections, while agency shapes which corrections are learned.

\paragraph{P5. Emergent Spatial Agency.}
The system develops spatially nonuniform responsiveness with $k_{\mathrm{eff}}$ mean $= 6.03$, std $= 0.73$, and range $[5.15,\,8.76]$. The partial correlation $\rho_{\mathrm{partial}}(k_{\mathrm{eff}}, H_{\mathrm{norm}} \mid n_{\mathrm{legal}}) = +0.179$ ($p < 10^{-4}$, $n = 1{,}002$) confirms that the modulation is structure-dependent rather than a trivial function of legal-move count alone. The learned responsiveness field is therefore not an engineered temperature schedule. It is an interaction-derived spatial modulation, exactly as predicted by the Agency claim.

\paragraph{P6. Transfer and Retention.}
Across three independent seeds, the chess domain yields positive backward transfer of $\mathrm{BT}_A = +35.4 \pm 2.9\cp$ and near-zero $\mathrm{BT}_B = +0.3 \pm 3.6\cp$. In the reference run, the full system reaches $\mathrm{BT}_A = +38.5\cp$ and $\mathrm{BT}_B = +0.4\cp$, exceeding Replay's $\mathrm{BT}_A = +26.8\cp$ without storing a single raw example. The Crucible is the mechanism behind that result: verified corrections carry useful structure across context boundaries, allowing transfer while preserving prior alignment.

\vspace{0.5em}
The developmental cascade is confirmed in the predicted order, as shown below:
\begin{center}
\scriptsize
\setlength{\tabcolsep}{3pt}
\begin{tabular}{@{}lcccc@{}}
\toprule
 & Full & No-Ag. & No-Cruc. & No-Cryst \\
\midrule
$\mathrm{BT}_A$ & $+38.5\cp$ & $+16.0\cp$ & $+15.6\cp$ & $-44.7\cp$ \\
\bottomrule
\end{tabular}
\end{center}

Crystallization provides persistence. Without it, backward transfer becomes strongly negative. The Crucible provides curation. Without it, corrections persist but cannot be curated for cross-context use. Agency drives the exploration that feeds the Crucible. Without agency, the Crucible encounters far fewer contradiction zones and therefore performs almost no selective self-correction.

The mechanistic evidence for that last claim is direct. In the full system, $7{,}344$ centers ever crystallize, $4{,}782$ are verified, and the Crucible processes $19{,}054$ dissolution events across the run. In the No-Agency training ablation, $7{,}190$ centers ever crystallize, $4{,}690$ are verified, and only $30$ dissolution events occur. The difference is not architectural but developmental: the same Crucible code is present in both runs, but only agency-guided exploration drives corrections into the contradiction zones that trigger selective curation. These are dissolution events, not unique centers; centers that dissolve, re-crystallize, and dissolve again are counted each time. The current reversal threshold therefore produces active churning, but the behavioral result shows the mechanism remains functional.

\subsection{Stage~III Results: Scalability Validation in CIFAR-100}
\label{sec:res:cifar}

Stage~III asks whether the same dynamics remain stable and useful in a standard benchmark at substantially higher dimensionality. All methods operate on the same frozen ViT-B/16 $+$ PCA representation and the same frozen coherence head, so the comparison tests correction dynamics on top of a shared evaluator rather than end-to-end representation learning. In this domain, faithful evaluation requires the log-space readout introduced in Section~\ref{sec:domains:cifar}, which aligns the correction field with the softmax-based baseline evaluator.

\paragraph{P7. Near-Zero Forgetting at Scale.}
IBF achieves $\mathrm{BT} = -0.004 \pm 0.000$ across $20$ sequential tasks at $68$ dimensions, preserving essentially all of the frozen coherence head's baseline performance while storing zero raw data. Replay loses $23.4\%$ under the same evaluator and with $5{,}000$ stored samples, while EWC and the vanilla MLP catastrophically forget. This is the central Stage~III result.

\begin{table}[htbp]
\centering
\caption{\textbf{CIFAR-100 results.} Task-IL values use the log-space readout described in Section~\ref{sec:domains:cifar}. IBF achieves replay-superior retention with zero stored raw data and retains a structural Class-IL advantage without a task oracle.}
\label{tab:cifar_results}
\vspace{4pt}
\footnotesize
\begin{tabular}{@{}lccc@{}}
\toprule
\textbf{Method} & \textbf{Task-IL Avg} & \textbf{BT} & \textbf{Class-IL} \\
\midrule
Full IBF        & $.892 \pm .009$ & $-.004 \pm .000$ & $.528$ \\
Replay MLP      & $.723$          & $-.234$          & $.392$ \\
EWC             & $.311$          & $-.681$          & $.052$ \\
MLP             & $.276$          & $-.719$          & $.049$ \\
Coherence head  & $.901$          & $+.000$          & --- \\
\bottomrule
\end{tabular}
\end{table}

\paragraph{A structural Class-IL advantage.}
Task-IL assumes task identity at test time; Class-IL does not, requiring prediction over all classes. IBF also retains a structural advantage under Class-IL evaluation: it reaches $52.8\%$ across all $100$ classes without a task oracle, whereas Replay reaches $39.2\%$ and the neural baselines collapse to near chance. This matters because the IBF readout natively evaluates the full class space through the corrected coherence landscape; it does not rely on an oracle to restrict the candidate set at test time.

\paragraph{P8. Regime-dependent agency.}
Agency is effectively neutral in CIFAR. Full IBF and No-Agency are indistinguishable to three decimal places in the representative seed ($0.903$ vs.\ $0.903$, $\mathrm{BT} = -0.004$ vs.\ $-0.003$). This is the correct null result. Claim~7 predicts that agency matters developmentally only when the trajectory itself provides room for it to matter, and CIFAR's fixed data stream does not. The same mechanism is mildly harmful in RRW, essential in chess, and neutral here.

\paragraph{Spatial isolation is the dominant mechanism at 68D.}
The CIFAR ablations show that kernel localization alone is sufficient for retention at this dimensionality. Removing the Crucible, crystallization, or agency changes almost nothing: all four conditions cluster around average accuracy $\approx 0.902$ and backward transfer $\approx -0.003$. In the representative seed, No-Crucible is even marginally best. The higher-order mechanisms still operate, with thousands of crystallized centers and tens of thousands of dissolutions, but at this scale they contribute little measurable behavior because inter-task distances are large relative to the calibrated bandwidth.

\paragraph{Corrections are real but ceiling-limited.}
The coherence head alone achieves $0.901$ Task-IL, and Full IBF averages $0.892 \pm 0.009$ across three seeds, effectively net-neutral against this strong prior. To test whether the correction layer still carries genuine signal, the setup was repeated with a weakened head trained on two samples per class. Under this weaker prior, IBF rises from $0.726$ to $0.739$ ($+1.3\%$), improving $16$ of $20$ tasks with a maximum single-task gain of $+8.2\%$. The corrections are real; the standard configuration simply leaves little room for measurable improvement.

\paragraph{Mechanism activation under a different regime.}
The behavioral redundancy of the higher-order mechanisms in CIFAR is itself a prediction of the theory: when inter-task distances are large relative to the calibrated bandwidth, spatial isolation alone is sufficient and the remaining mechanisms contribute nothing measurable. A contrasting regime is now beginning to emerge in the frozen-LLM extension introduced in Section~\ref{sec:domains:cifar}. There, where the correction layer must carry much more of the behavioral load, the same mechanisms begin to separate: removing crystallization produces severe forgetting, removing the Crucible preserves retention but allows correction amplitudes to grow unchecked, and removing agency remains effectively neutral, as in CIFAR. The details belong to a companion write-up. What matters here is the contrast in regime: the same mechanisms that are behaviorally redundant at $68$D in CIFAR become non-redundant once spatial isolation alone is no longer enough.

\subsection{Cross-Domain Pattern}
\label{sec:res:pattern}

Taken together, the three validation domains reveal a functional hierarchy of mechanisms. Spatial isolation is foundational and scales across all three domains. It is sufficient for strong retention in CIFAR, where tasks are naturally well separated in a high-dimensional latent space. Crystallization becomes important in RRW, where memories must survive explicit phase transitions. Agency and the Crucible become decisive in chess, where the system must discover, verify, and curate shared strategic structure rather than merely isolate contradictory or disjoint corrections.

The cross-domain regime dependence of agency is especially important. In RRW, agency slightly increases forgetting because exploration drives corrections into maximally contradictory cross-context territory. In chess, the same mechanism becomes essential: removing it sharply reduces backward transfer and collapses much of the behavioral advantage. In CIFAR, its effect becomes negligible because the domain provides little trajectory dependence for responsiveness to modulate. The mechanism is unchanged; the environment determines whether it is harmful, useful, or effectively irrelevant.

The same staged reading applies to the Crucible. In RRW it is transfer-enabling but low-yield, because there is almost no universal structure to promote. In chess it becomes a genuine self-correction system, verifying reusable corrections and dissolving contradicted ones before they can be redeployed across contexts. In CIFAR it is behaviorally redundant because geometric separation already prevents interference. This pattern is not ad hoc. It is the one the framework predicts: higher-order mechanisms become active only when the environment provides the conditions they require.

Finally, the geometric calibration story remains coherent across all three domains. RRW, chess, and CIFAR all admit a usable training bandwidth derived from latent geometry rather than grid search, and the resulting empirical scaling ratios remain of the same order of magnitude despite the jump from $8$ to $68$ dimensions.

\section{Discussion}
\label{sec:discussion}

The Results section answered the experimental program stage by stage. The role of the Discussion is to interpret what those confirmations establish. 

We begin by returning to the paper's premise, place it against adjacent frameworks, and finally state the scope limits and next empirical frontiers as directly as possible.

\subsection{The Alignment Premise Under Test}
\label{sec:disc:demonstrated}

This paper opened with a premise: information is not data, but the achievement of structural alignment between a system's internal configuration and the structure of its environment. The entire continual-learning program, from the toy model through the claims, engine, staged experiments, and results, was a test of whether that premise generates correct mechanistic consequences. The question now is not whether the premise is philosophically appealing, but whether it generated mechanisms that behaved as predicted under empirical stress.

\paragraph{Memory as preserved alignment.}
The central claim of the paper is that memory need not be stored as globally shared parameter superposition. Under the Modification Dynamics, memory becomes persistent local deformation of the effective coherence landscape. The three domains support that deduction at increasing levels of complexity. In RRW, IBF reduces forgetting by $43\%$ relative to Replay MLP without replaying raw observations. In chess, the result is stronger and remains positive across three independent seeds: $\mathrm{BT}_A = +35.4 \pm 2.9\cp$, exceeding Replay's $+26.8\cp$ without storing a single past position. In CIFAR-100, the same dynamics preserve performance across $20$ sequential tasks with near-zero forgetting ($\mathrm{BT} = -0.004$). These are not three disconnected results. They are three versions of the same proposition: retention can arise from persistent local structure rather than from replay, regularization, or frozen parameter partitions.

\paragraph{Agency as a regime-dependent developmental mechanism.}
The Agency claim predicts something stronger than confidence or temperature. It predicts that responsiveness emerges from interaction as a spatially varying developmental force, and that its contribution depends on the discrepancy regime. The cross-domain pattern supports that prediction closely. In RRW, agency is mildly harmful because exploration pushes corrections into maximally contradictory cross-context territory. In chess, agency becomes consequential during development: in the reference run at the geometrically prescribed bandwidth, removing it reduces the behavioral advantage from $+90.2\cp$ to $+81.9\cp$, lowers $\mathrm{BT}_A$ from $+38.5\cp$ to $+16.0\cp$, and collapses Crucible dissolution activity from $19{,}054$ events to just $30$. In CIFAR, its effect becomes negligible because the task stream offers little trajectory dependence for responsiveness to modulate. The mechanism is unchanged; what changes is whether the environment gives it anything useful to do.

\paragraph{Intelligence as curated memory under agency.}
The Intelligence claim is not that memory alone improves behavior. It is that memory and agency together can shape a correction landscape that yields systematically higher-alignment outcomes. Chess provides the clearest test. At the geometrically prescribed bandwidth, IBF achieves a mean advantage of $+88.9 \pm 2.8\cp$ over the passive baseline across three seeds, with the reference run reaching $+90.2\cp$. The No-Agency training ablation shows why: agency influences which corrections are learned, and therefore whether the resulting landscape is merely competent or strategically strong.

\paragraph{Self-correction through contradiction.}
The Crucible is not what protects memory. Gating already does that, as RRW's No-Crucible condition makes clear. Its role is to curate what can survive cross-context use. In RRW, where the phases are maximally contradictory, that role is visible but low-transfer: there is little genuinely universal structure to preserve across contexts. In chess, the same mechanism becomes decisive. With agency, 7,344 centers ever crystallize, 4,782 are verified, and the Crucible processes 19,054 dissolution events; without agency, 7,190 centers crystallize, 4,690 are verified, and only 30 dissolution events occur. Agency therefore matters not by adding memory, but by driving the system into the contradictory regions where selective dissolution can occur. In CIFAR, the Crucible still operates, but adds little measurable behavior because spatial isolation already prevents meaningful interference.

Taken together, the premise generated a concrete architecture; the architecture generated falsifiable predictions; and the predictions survived controlled ablation, independent evaluation, and cross-domain transfer. Within the scope of the present program, the theory did not merely remain consistent with the data. It generated mechanisms whose predicted signatures the data then confirmed.

\subsection{Relation to Existing Frameworks}
\label{sec:disc:related}

To situate IBF properly, it helps to compare it not first to methods, but to frameworks that try to explain adaptive organization at the level of principle. In that sense, the closest existing relative is Friston's \emph{Free Energy Principle}~\cite{friston2010free}. Both begin from the intuition that persistent adaptive systems survive by systematically reducing \emph{mismatch with their environment}. But the resemblance is mostly one of ambition. FEP begins from \emph{Bayesian inference}: the system is understood through a generative model, and behavior is derived as approximate posterior updating under that model. IBF begins from \emph{coherence dynamics on configuration space}. No generative model is assumed at the outset; what changes through interaction is the \emph{effective coherence landscape itself}, deformed locally by discrepancy. From that difference follows another one: in IBF, \emph{memory, agency, and self-correction emerge as distinct consequences of the dynamics}, with distinct empirical signatures, rather than as different aspects of a single inferential process.

At a more immediate architectural level, IBF also stands near traditions that discovered, each in its own way, that \emph{stable learning cannot be built entirely on global interference}. ART treated the stability--plasticity dilemma as a structural problem~\cite{grossberg1987competitive, carpenter1991artmap}. Episodic-control methods showed that kernel-weighted retrieval of stored experiences can bypass gradient interference~\cite{blundell2016model, pritzel2017neural}. Kernel and localist methods rely on the same broad geometric intuition: if memory is to survive continued learning, corrections must remain \emph{local rather than globally smeared across parameter space}~\cite{titsias2020functional, derakhshani2021kernel, pan2021continual, buhmann2003radial}. IBF clearly belongs in that neighborhood. But ART stores categories, episodic methods store traces, and kernel methods store support points. IBF stores \emph{modification sites whose thermodynamic state changes through interaction}: they crystallize when local discrepancy converges, remain silent across contexts until evidence verifies them, dissolve when later evidence turns against them, and shape future responsiveness through the same law that created them.

The sharpest contrast remains with the dominant continual-learning paradigm itself. Elastic Weight Consolidation protects important parameters~\cite{kirkpatrick2017overcoming}. Progressive and dynamically expandable networks preserve prior competence by freezing or growing architecture~\cite{rusu2016progressive, yoon2018den}. Replay methods keep old tasks statistically present through stored observations~\cite{rolnick2019experience}. Context-dependent gating protects prior structure through externally supplied task masks~\cite{masse2018alleviating}. These methods can work well, and some serve as baselines in the present paper. But they all address forgetting inside a substrate where memory still resides in \emph{globally shared parameters}. IBF changes that condition itself. \emph{Memory becomes persistent local deformation rather than protected weight configuration.} \emph{Gating becomes a state of the memory rather than an external routing instruction.} The agency channel that shapes which corrections are learned arises from the same modification dynamics that produce locality in the first place. That is why IBF is best read not as one more improvement inside the existing paradigm, but as \emph{an attempt to formulate a different substrate beneath it.}

\subsection{Limitations}
\label{sec:disc:limitations}

The clearest limit of the present paper is also the most important to state plainly: \emph{this is not yet an end-to-end account of representation formation from raw interaction}. The framework is validated here on top of a supplied encoder and a supplied baseline evaluator. The encoder provides the configuration space; the evaluator provides the initial coherence landscape; IBF contributes the modification dynamics that reshape that landscape over time. That is not a hidden assumption, and it is not a minor detail. It is the explicit scope of the present empirical program.

\emph{A second limit is one of scale}. The validations reported here span latent spaces from $8$ to $68$ dimensions, with the largest effective dimensionality at $d_{\mathrm{eff}} = 28.9$. That is enough to test the substrate claim in a meaningful way, but it remains far from the geometry of frontier foundation-model latents. CIFAR shows that the foundational retention layer remains stable at this scale. It does not yet show how the full mechanism stack behaves when the representational regime becomes vastly larger and denser. A first step beyond this range is already emerging in the frozen-LLM instantiation discussed briefly, but that result belongs to a separate empirical treatment and is not part of the present validation set.

The evidential \emph{role of the three domains is also not symmetrical}. Chess is the cleanest test of the framework's behavioral claim, because evaluation is performed by a stronger external oracle on disjoint positions. CIFAR serves a different purpose. It is primarily a scaling and non-destructiveness test. Because the coherence head is trained on the same label space as the evaluation metric, the CIFAR result does not support the same kind of independence claim as chess. Its evidential role is different, not weaker.

There is \emph{a further limit in how discrepancy is supplied}. In RRW, the environment is analytically defined. In chess, discrepancy is anchored to Stockfish during training. In CIFAR, it is derived from supervised class-conditioned structure. These settings are sufficient for the staged validation carried out here, but they are still partially scaffolded. Real environments do not announce their phase boundaries, and they do not provide a trusted external arbiter of contradiction. The present engine therefore demonstrates continual correction, preservation, and selective dissolution under controlled conditions, but not yet autonomous continual learning in an open stream.

Finally, \emph{the paper remains a single-agent study}. It demonstrates memory, agency, and self-correction for one system interacting with an environment. It does not yet test the broader communication and multi-agent arc of the framework, in which agents mutually deform one another's effective landscapes. Nor does it say much about environments with no stable discrepancy structure at all. In pure noise, nothing meaningful crystallizes. That is simply outside the range of conditions this paper tests.

These limitations narrow the scope of the present results. They do not make them less important. They simply locate the paper where it actually stands: as a substrate-level proposal that has been made formal enough to test, and tested far enough to justify stronger questions.

\subsection{Future Directions}
\label{sec:disc:future}

The empirical engine presented in this paper implements only the minimal discrete mechanics required to test the continual-learning arc of the framework. The next steps are the next places where the premise becomes vulnerable in a useful way.

\emph{The most immediate frontier is scale.} CIFAR-100 shows that the foundational layer, spatial isolation, remains stable up to $68$ dimensions and can preserve a strong baseline across long task sequences. The natural next step is to move into much larger frozen latent spaces, on the order of hundreds or thousands of effective dimensions, and ask not only whether the dynamics still function there, but when the higher-order mechanisms become behaviorally necessary. If geometric overlap increases, do verification, dissolution, and responsiveness modulation become indispensable rather than merely available? A first indication that this scaling question may already open beyond the present range is emerging in the frozen-LLM instantiation mentioned. 

\emph{A second frontier is continuity.} The current Crucible still operates in a setting with explicit context transitions and phase-local resets. A stricter test would be a boundary-free stream in which memories must crystallize, fall silent, reactivate, and dissolve under rolling discrepancy statistics alone. That would bring the implementation closer to the spirit of the theory, which is continuous-time and thermodynamic rather than phase-scripted.

\emph{A third frontier is multi-agent interaction.} In all three present domains, discrepancy is supplied by an external environment or oracle. A more demanding setting would place multiple IBF agents in a shared world where part of each agent's discrepancy structure is generated by the others. That would turn communication and mutual deformation from a theoretical promise into an empirical question, and would show whether decentralized coordination can emerge without a single externally imposed objective.

\emph{The framework should also be tested in richer world-model domains.} Chess already shows that the mechanism can discover and curate structured strategic regularities, while CIFAR shows that the foundational retention layer remains stable in a larger representational regime. The next step is to move into continuous-control, robotics, or simulated physical environments where action-conditioned latent world models matter directly. 

That is also the setting in which the relation between IBF and the broader post-scaling transition becomes most concrete. A first indication that this route may extend beyond the present validation domains is already emerging in companion work: \emph{the same IBF engine is being applied on top of a frozen 7B language model in a proposition-space setting}, where it appears able to inject fictional knowledge, preserve it through later learning, and selectively revise it under contradiction without modifying the base model's parameters.

Finally, \emph{there is the question of substrate in the literal sense}. IBF is conceptually a poor match for dense globally synchronous computation and a much better match for local, event-driven, asynchronous processes. Its primitives are localized nucleation, passive decay, stability transitions, and sparse interaction among nearby memory sites. That makes neuromorphic or analog implementations more than a hardware curiosity. If the substrate claim of the paper is right, such environments may not merely run IBF faster. They may be the places where it becomes most natural.

\vspace{0.5em}
\begin{center}
\emph{---}
\end{center}

\noindent
To conclude, the central claim of this paper is that continual learning does not need to be built on destructive superposition. Across three domains of increasing complexity, the same two-equation substrate produced persistent memory, selective self-correction, and regime-dependent agency without replaying raw experience. Within the scope of the present validation program, the premise produced the mechanisms, the mechanisms produced the predictions, and the predictions survived empirical testing. \textbf{That is the result.}

\vspace{1em}

\section*{Reproducibility Release}

The reference implementation accompanying this paper is available at:

\begin{center}
\url{https://github.com/negulescu42/information-as-alignment}
\end{center}

The repository contains the toy model, the three validation domains (RRW, chess, and CIFAR-100), the JSON result files corresponding to the reported runs, and a short instruction file describing how to inspect or reproduce the main experiments.

\bibliographystyle{unsrtnat}

\begin{thebibliography}{17}
\providecommand{\natexlab}[1]{#1}
\providecommand{\url}[1]{\texttt{#1}}
\expandafter\ifx\csname urlstyle\endcsname\relax
  \providecommand{\doi}[1]{doi: #1}\else
  \providecommand{\doi}{doi: \begingroup \urlstyle{rm}\Url}\fi

\bibitem[McCloskey and Cohen(1989)]{mccloskey1989catastrophic}
Michael McCloskey and Neal~J. Cohen.
\newblock Catastrophic interference in connectionist networks: The sequential learning problem.
\newblock \emph{Psychology of Learning and Motivation}, 24:\penalty0 109--165, 1989.

\bibitem[French(1999)]{french1999catastrophic}
Robert~M. French.
\newblock Catastrophic forgetting in connectionist networks.
\newblock \emph{Trends in Cognitive Sciences}, 3\penalty0 (4):\penalty0 128--135, 1999.

\bibitem[Kirkpatrick et~al.(2017)Kirkpatrick, Pascanu, Rabinowitz, Veness, Desjardins, Rusu, Milan, Quan, Ramalho, Grabska-Barwinska, et~al.]{kirkpatrick2017overcoming}
James Kirkpatrick, Razvan Pascanu, Neil Rabinowitz, Joel Veness, Guillaume Desjardins, Andrei~A. Rusu, Kieran Milan, John Quan, Tiago Ramalho, Agnieszka Grabska-Barwinska, et~al.
\newblock Overcoming catastrophic forgetting in neural networks.
\newblock \emph{Proceedings of the National Academy of Sciences}, 114\penalty0 (13):\penalty0 3521--3526, 2017.

\bibitem[Rolnick et~al.(2019)Rolnick, Ahuja, Schwarz, Lillicrap, and Wayne]{rolnick2019experience}
David Rolnick, Arun Ahuja, Jonathan Schwarz, Timothy~P. Lillicrap, and Gregory Wayne.
\newblock Experience replay for continual learning.
\newblock In \emph{Advances in Neural Information Processing Systems}, volume~32, 2019.

\bibitem[Yoon et~al.(2018)Yoon, Yang, Lee, and Hwang]{yoon2018den}
Jaehong Yoon, Eunho Yang, Jeongtae Lee, and Sung~Ju Hwang.
\newblock Lifelong learning with dynamically expandable networks.
\newblock In \emph{International Conference on Learning Representations (ICLR)}, 2018.

\bibitem[Rusu et~al.(2016)Rusu, Rabinowitz, Desjardins, Soyer, Kirkpatrick, Kavukcuoglu, Pascanu, and Hadsell]{rusu2016progressive}
Andrei~A. Rusu, Neil~C. Rabinowitz, Guillaume Desjardins, Hubert Soyer, James Kirkpatrick, Koray Kavukcuoglu, Razvan Pascanu, and Raia Hadsell.
\newblock Progressive neural networks.
\newblock In \emph{arXiv preprint arXiv:1606.04671}, 2016.

\bibitem[McCulloch and Pitts(1943)]{mcculloch1943logical}
Warren~S. McCulloch and Walter Pitts.
\newblock A logical calculus of the ideas immanent in nervous activity.
\newblock \emph{The Bulletin of Mathematical Biophysics}, 5\penalty0 (4):\penalty0 115--133, 1943.
\newblock \doi{10.1007/BF02478259}.

\bibitem[Friston(2010)]{friston2010free}
Karl Friston.
\newblock The free-energy principle: a unified brain theory?
\newblock \emph{Nature Reviews Neuroscience}, 11\penalty0 (2):\penalty0 127--138, 2010.

\bibitem[Grossberg(1987)]{grossberg1987competitive}
Stephen Grossberg.
\newblock Competitive learning: From interactive activation to adaptive resonance.
\newblock \emph{Cognitive Science}, 11\penalty0 (1):\penalty0 23--63, 1987.

\bibitem[Carpenter et~al.(1991)Carpenter, Grossberg, and Reynolds]{carpenter1991artmap}
Gail~A. Carpenter, Stephen Grossberg, and John~H. Reynolds.
\newblock {ARTMAP}: Supervised real-time learning and classification of nonstationary data by a self-organizing neural network.
\newblock \emph{Neural Networks}, 4\penalty0 (5):\penalty0 565--588, 1991.

\bibitem[Blundell et~al.(2016)Blundell, Uria, Pritzel, Li, Ruderman, Leibo, Rae, Wierstra, and Hassabis]{blundell2016model}
Charles Blundell, Benigno Uria, Alexander Pritzel, Yazhe Li, Avraham Ruderman, Joel~Z. Leibo, Jack Rae, Daan Wierstra, and Demis Hassabis.
\newblock Model-free episodic control.
\newblock \emph{arXiv preprint arXiv:1606.04460}, 2016.

\bibitem[Pritzel et~al.(2017)Pritzel, Uria, Srinivasan, Puigdom{\`e}nech, Vinyals, Hassabis, Wierstra, and Blundell]{pritzel2017neural}
Alexander Pritzel, Benigno Uria, Sriram Srinivasan, Adri{\`a} Puigdom{\`e}nech, Oriol Vinyals, Demis Hassabis, Daan Wierstra, and Charles Blundell.
\newblock Neural episodic control.
\newblock In \emph{International Conference on Machine Learning}, 2017.

\bibitem[Titsias et~al.(2020)Titsias, Schwarz, Matthews, Pascanu, and Teh]{titsias2020functional}
Michalis~K. Titsias, Jonathan Schwarz, Alexander G. de~G. Matthews, Razvan Pascanu, and Yee~Whye Teh.
\newblock Functional regularisation for continual learning with gaussian processes.
\newblock In \emph{International Conference on Learning Representations}, 2020.

\bibitem[Derakhshani et~al.(2021)Derakhshani, Zhen, Shao, and Snoek]{derakhshani2021kernel}
Mohammad~Mahdi Derakhshani, Xiantong Zhen, Ling Shao, and Cees Snoek.
\newblock Kernel continual learning.
\newblock In \emph{International Conference on Machine Learning}, 2021.

\bibitem[Pan et~al.(2021)Pan, Swaroop, Immer, Eschenhagen, Turner, and Khan]{pan2021continual}
Pingbo Pan, Siddharth Swaroop, Alexander Immer, Runa Eschenhagen, Richard~E. Turner, and Mohammad~Emtiyaz Khan.
\newblock Continual deep learning by functional regularisation of memorable past.
\newblock In \emph{Advances in Neural Information Processing Systems}, 2021.

\bibitem[Buhmann(2003)]{buhmann2003radial}
Martin~D. Buhmann.
\newblock \emph{Radial Basis Functions: Theory and Implementations}.
\newblock Cambridge University Press, 2003.

\bibitem[Masse et~al.(2018)Masse, Grant, and Freedman]{masse2018alleviating}
Nicolas~Y. Masse, Gregory~D. Grant, and David~J. Freedman.
\newblock Alleviating catastrophic forgetting using context-dependent gating and synaptic stabilization.
\newblock \emph{Proceedings of the National Academy of Sciences}, 115\penalty0 (44):\penalty0 E10467--E10475, 2018.

\end{thebibliography}

 
\begin{table*}[p]
\centering
\caption{\textbf{Formal Primitives of the Informational Buildup Framework.}
  Nine primitives define the ontological substrate.  The complete
  formal treatment, including structural properties and inter-primitive
  relations, is provided in a dedicated foundational work.}
\label{tab:primitives}
\vspace{4pt}
\small
\begin{tabular}{@{}p{2.8cm}p{1.8cm}p{8.5cm}@{}}
\toprule
\textbf{Primitive} & \textbf{Symbol} & \textbf{Definition} \\
\midrule
Field &
$\Field$ &
A nonempty universal set whose elements are informational
patterns; the total space of possible informational configurations. \\[4pt]
Pattern &
$\psi \in \Field$ &
An element of the Field.  Patterns possess no intrinsic meaning
or stability independent of their relations to systems and scales. \\[4pt]
Closed Informational System (CIS) &
$S \subset \Field$ &
A connected region of configuration space where coherence exceeds
a viability threshold and boundary gradients point inward,
actively resisting dissolution. \\[4pt]
Observational Scale &
$\lambda \in \Lambda$ &
A resolution parameter specifying the level of abstraction at
which patterns and systems are distinguished.  Determines which
distinctions are operational and which coherence function applies. \\[4pt]
Configuration Space &
$\ConfigSpace$ &
For each system $S$ at scale $\lambda$, the differentiable
manifold of all admissible internal states.  Every point
represents a possible internal configuration. \\[4pt]
Coherence &
$\Coh(x_S, S, \lambda)$ &
A scalar function quantifying the degree of structural alignment
between a system's internal configuration and its environment.
Relational, fragile, non-reducible. \\[4pt]
Informational Gravity &
$G(x_S) = \nabla_{x_S}\Coh$ &
The gradient of coherence on configuration space: the pull toward
higher structural alignment. \\[4pt]
Responsiveness &
$k_S(x_S) > 0$ &
A positive-valued function quantifying a system's capacity to
follow coherence gradients.  Modulated by interaction history
(the agency channel). \\[4pt]
Process Time &
$t$ &
A parameter indexing the sequence of system reconfigurations
driven by coherence dynamics. \\
\bottomrule
\end{tabular}
\end{table*}
 
\begin{table*}[p]
\centering
\caption{\textbf{Formal Structure and Placement in the Broader Arc.}
The four axioms and the local modification postulate stated in this paper define the minimal structure needed to ground the continual learning experimental program. The table also situates that local arc within the broader foundational framework, whose remaining closure postulates and claims are shown here in condensed form. Rows marked \textit{Here} are stated or empirically instantiated in this paper; rows marked \textit{Foundational} belong to the larger dedicated treatment. Claims 9 and 10 are operationalized jointly in this paper as the \textit{Self-Correction} capacity: the system reflexively tracks its cross-context coherence state (Thm.~9) to selectively trigger the thermodynamic dissolution of false alignment (Thm.~10).}
\label{tab:formal_summary}
\vspace{4pt}

\scriptsize
\setlength{\tabcolsep}{5pt}
\renewcommand{\arraystretch}{1.15}

\begin{tabular*}{\textwidth}{@{\extracolsep{\fill}}lp{3.2cm}p{9.3cm}l@{}}
\toprule
\textbf{Ref.} & \textbf{Name} & \textbf{Statement / Role} & \textbf{Scope} \\
\midrule
\multicolumn{4}{@{}l}{\textit{Axioms (Existence and Motion)}} \\
\midrule
I & The Field & $\Field \neq \emptyset$; for each $(S,\lambda)$, the Field admits an associated differentiable configuration space $\ConfigSpace$. & Here \\
II & Coherence & $\Coh: \ConfigSpace \to \mathbb{R}^{+}$; relational, fragile, and non-decomposable. Defines viability and CIS boundaries. & Here \\
III & Informational Gravity & $G = \nabla_{x_S}\Coh$; systems experience a pull toward higher structural alignment. & Here \\
IV & Law of Motion & $\dot{x}_S = k_S \cdot \nabla_{x_S}\Coheff$; motion follows the effective coherence gradient, modulated by responsiveness. & Here \\[6pt]

\multicolumn{4}{@{}l}{\textit{Closure Postulates}} \\
\midrule
0 & Scale--Coherence Consist. & Legitimizes representing coherence dynamics on configuration space once observational scale is fixed. & Foundational \\
I & System Interaction & Interaction forms a joint coherence landscape and creates the discrepancy conditions for system influence. & Foundational \\
II & Recursive Scale Struct. & Relates system realization across scales and permits upward propagation of modification. & Foundational \\
III & Dynamic Stability & Specifies stability conditions for persistence, basin loss, and renewed organization after destabilization. & Foundational \\
IV & Modification Dynamics & $\frac{\partial}{\partial t} \dCoh = \eta\,\Kern\,\Dscr - \mu\,\dCoh$; localized landscape deformation driven by discrepancy signals. & Here (trimmed) \\[6pt]

\multicolumn{4}{@{}l}{\textit{Claim Arc}} \\
\midrule
1 & Emergence \& Persistence & Viable coherence basins yield Closed Informational Systems that arise, stabilize, and persist. & Foundational \\
2 & Identity & System identity is given by invariance of the same coherence basin across admissible perturbation. & Foundational \\
3 & Memory & Persistent landscape modification yields preserved alignment as long-lived deformation of the landscape. & Here / Found. \\
4 & Communication & Interaction deforms another system's effective gradient field, creating the conditions for signal transfer. & Foundational \\
5 & Learning & Repeated discrepancy accumulates modification and alters future trajectories. & Foundational \\
6 & Replication & Persistent internal constraints can be externalized through interaction, producing a new coherence basin. & Foundational \\
7 & Agency & Nontrivial interaction produces spatially nonuniform responsiveness, modulating how gradients are followed. & Here / Found. \\
8 & Intelligence & Memory and agency jointly yield systematically higher-alignment behavior than the unmodified baseline. & Here / Found. \\
9 & Reflexive Coherence \newline (Self-Correction) & A system tracking its coherence state can reflexively regulate dynamics. Instantiated here via cross-context memory verification. & Here / Found. \\
10 & Dissolution \newline (Entropic Return) & Loss of viable basins returns organized structure to the Field. Instantiated here via the thermodynamic melting of memory loci ($\mu \to \mu_{\mathrm{base}}$). & Here / Found. \\
11 & Discrete Convergence & Under Conditions R, R$'$, and A, the discrete implementation converges to the continuous dynamics. & Here / Found. \\
\bottomrule
\end{tabular*}
\end{table*}

\begin{table*}[p]
\centering
\caption{\textbf{Claim-to-Engine Mapping.}
Domain-agnostic correspondence between the formal theory (Section~\ref{sec:theory}) and the universal discrete engine (Section~\ref{sec:engine}). The table records how each theoretical construct is realized at the engine level. Domain-specific architectures, losses, and evaluation protocols enter only later, in Section~\ref{sec:domains}.}
\label{tab:engine_mapping}
\vspace{4pt}

\scriptsize
\setlength{\tabcolsep}{4pt}
\renewcommand{\arraystretch}{1.08}

\begin{tabular*}{\textwidth}{@{\extracolsep{\fill}}p{4.0cm}p{7.4cm}p{4.0cm}@{}}
\toprule
\textbf{Theoretical Construct} & \textbf{Engine Realization} & \textbf{Local Consequence} \\
\midrule
\multicolumn{3}{@{}l}{\textit{Axioms (Existence and Motion)}} \\[3pt]

Field $\Field$ (Axiom~I) & Domain observation space together with the frozen encoder $\pi$ and induced latent configuration space $z \in \mathbb{R}^{d}$ & The engine always operates on admissible latent configurations \\[3pt]

Coherence $\Coh$ (Axiom~II) & Frozen baseline evaluator $\hat{\Coh}_{\mathrm{base}}(z)$ plus effective correction field $\delta\hat{R}(z)$ & Effective coherence landscape $\Coheff(z) = \hat{\Coh}_{\mathrm{base}}(z) + \delta\hat{R}(z)$ \\[3pt]

Informational Gravity $G = \nabla_{x_S}\Coh$ (Axiom~III) & Discrete coherence increments over candidate next states, $\Delta R^{\mathrm{eff}}_j$ & Local directional preference toward higher alignment \\[3pt]

Law of Motion (Axiom~IV) & Boltzmann action selection with responsiveness $\keff(z)$ over candidate next states & Probabilistic ascent on the effective coherence landscape \\[6pt]

\multicolumn{3}{@{}l}{\textit{Modification Postulate}} \\[3pt]

Discrepancy signal $\Dscr$ & Interaction-derived gap between imposed coherence structure and current effective evaluation & Driver of local modification; distinct from the motion gradient \\[3pt]

Localization kernel $\Kern$ & Gaussian radial basis kernel with per-particle bandwidth $\sigma_i$ & Spatially local readout and writing \\[3pt]

Modification dynamics $\partial_t \dCoh = \eta\,\Kern\,\Dscr - \mu\,\dCoh$ & Kernel-weighted particle updates in the coherence-correction population & Local landscape deformation and memory formation \\[3pt]

Parallel responsiveness modification $\delta k_S$ & Variance-sensitive updates in the responsiveness population & Spatially nonuniform commitment / caution \\[3pt]

Crystallization ($\mu \to 0$) & Stability transition of particles whose recent discrepancy history converges & Long-lived local structure \\[3pt]

Dissolution ($\mu \to \mu_{\mathrm{base}}$) & Crucible-triggered loss of stability under sustained contradiction & Selective removal of false stability \\[3pt]

Capacity constraint & Merge-and-retain policy under finite particle budget & Bounded local resolution and memory density control \\[6pt]

\multicolumn{3}{@{}l}{\textit{Derived Capacities in the Engine}} \\[3pt]

Memory & Persistent coherence-correction particles with low effective decay & Retention without replay of past observations \\[3pt]

Agency & Responsiveness field $\keff(z)=\max(k_{\min},\,k_0+\delta k(z))$ & Spatially differentiated exploration and commitment \\[3pt]

Intelligence & Joint action of memory-shaped landscapes and agency-modulated motion & Systematically improved behavioral deployment of alignment \\[3pt]

Self-Correction & Reflexive read-gating, cross-context verification, and Crucible dissolution & Selective withdrawal of invalidated modifications \\[3pt]

Discrete Convergence & Finite particles, finite action sets, and latent evaluation under Conditions~R, R$'$, and~A & Faithful discrete approximation of the continuous dynamics \\
\bottomrule
\end{tabular*}
\end{table*}

\begin{table*}[p]
\centering
\caption{\textbf{Engine Symbol Dictionary I: Latent Space and Particle State.}
Core symbols defining the latent configuration space, effective evaluation, and particle-level state used by the universal IBF engine.}
\label{tab:engine_dictionary_1}
\vspace{4pt}

\scriptsize
\setlength{\tabcolsep}{4pt}
\renewcommand{\arraystretch}{1.08}

\begin{tabular*}{\textwidth}{@{\extracolsep{\fill}}p{1.9cm}p{4.5cm}p{3.2cm}p{5.3cm}@{}}
\toprule
\textbf{Symbol} & \textbf{Definition} & \textbf{Engine Realization} & \textbf{Notes} \\
\midrule
\multicolumn{4}{@{}l}{\textit{Latent Configuration and Evaluation}} \\[3pt]

$z$ & Current latent configuration & Frozen encoder output $\pi(\omega)$ & The engine operates only in latent space \\[3pt]

$z_j$ & Candidate next latent state & Latent image of action $a_j$ & Used for discrete motion / action selection \\[3pt]

$\hat{\Coh}_{\mathrm{base}}(z)$ & Baseline coherence evaluator & Frozen domain-specific baseline & No backpropagation through engine state \\[3pt]

$\delta\hat{R}(z)$ & Coherence-correction field & Kernel readout from coherence particles & Additive modification of the baseline evaluator \\[3pt]

$\Coheff(z)$ & Effective coherence & $\hat{\Coh}_{\mathrm{base}}(z) + \delta\hat{R}(z)$ & Read path output for motion \\[3pt]

$s_j^{\mathrm{eff}}$ & Effective coherence score of candidate state $z_j$ & $\Coheff(z_j)$ & Used in Boltzmann selection over candidate next states \\[3pt]

$\Delta R^{\mathrm{eff}}_j$ & Effective coherence increment & $\Coheff(z_j)-\Coheff(z_{\mathrm{current}})$ & Discrete analogue of gradient ascent \\[6pt]

\multicolumn{4}{@{}l}{\textit{Particle State}} \\[3pt]

$c_i$ & Generic particle & $(z_i, a_i, \sigma_i, \mu_{\mathrm{eff},i}, \mathrm{ctx}_i,\ldots)$ & Shared notation across both particle populations \\[3pt]

$z_i$ & Particle location & Latent center & Center of local support \\[3pt]

$a_i$ & Generic particle amplitude & $v_i$ (coherence channel) or $w_i$ (responsiveness channel) & Channel-specific payload \\[3pt]

$v_i$ & Coherence-correction amplitude & Stored correction weight & Contributes additively to $\delta\hat{R}(z)$ \\[3pt]

$w_i$ & Responsiveness amplitude & Stored responsiveness weight & Contributes intensively to $\delta k(z)$ \\[3pt]

$\sigma_i$ & Particle bandwidth & Per-particle kernel width & Kernel bandwidth; evaluated at the prescribed geometric scale \\[3pt]

$\mu_{\mathrm{eff},i}$ & Effective decay rate & Transient or crystallized regime & Controls passive fading \\[3pt]

$\mathrm{ctx}_i$ & Birth context & Context label stored with particle & Used by reflexive read-gating \\[3pt]

verified & Cross-context broadcast status & Verification flag for crystallized particles & Governs cross-context readability \\
\bottomrule
\end{tabular*}
\end{table*}

\begin{table*}[p]
\centering
\caption{\textbf{Engine Symbol Dictionary II: Read Path, Write Path, and Lifecycle.}
Core operational symbols governing kernel readout, localized writing, contradiction testing, and thermodynamic state transitions in the universal IBF engine.}
\label{tab:engine_dictionary_2}
\vspace{4pt}

\scriptsize
\setlength{\tabcolsep}{4pt}
\renewcommand{\arraystretch}{1.08}

\begin{tabular*}{\textwidth}{@{\extracolsep{\fill}}p{1.9cm}p{4.5cm}p{3.2cm}p{5.3cm}@{}}
\toprule
\textbf{Symbol} & \textbf{Definition} & \textbf{Engine Realization} & \textbf{Notes} \\
\midrule
\multicolumn{4}{@{}l}{\textit{Read Path}} \\[3pt]

$\Kern(z,z_i)$ & Localization kernel & Gaussian RBF $\exp(-\|z-z_i\|^2 / 2\sigma_i^2)$ & Shared by read and write paths \\[3pt]

$\gamma_i$ & Read-gating variable & Same-context or verified cross-context read access & $\gamma_i \in \{0,1\}$ \\[3pt]

$\delta k(z)$ & Responsiveness correction field & Intensive readout from crystallized responsiveness particles & Bounded local modulation \\[3pt]

$\keff(z)$ & Effective responsiveness & $\max(k_{\min},\,k_0+\delta k(z))$ & Controls Boltzmann sharpness \\[3pt]

$P(a_j\mid z_{\mathrm{current}})$ & Action-selection probability & Boltzmann distribution over candidate next states & Discrete instantiation of the Law of Motion \\[6pt]

\multicolumn{4}{@{}l}{\textit{Write Path and Lifecycle}} \\[3pt]

$\Dscr$ & Discrepancy signal & Gap between imposed coherence structure and current effective evaluation & Drives modification, not motion \\[3pt]

$\Dscr \cdot \Kern$ & Kernel-local discrepancy & Same-context spatial learning signal & Written into local particle histories \\[3pt]

$\bar{\Dscr}_{\mathrm{recent}}$ & Mean recent local discrepancy & Recent history statistic for convergence & Used for crystallization tests \\[3pt]

$\bar{\Dscr}_{\mathrm{raw,recent}}$ & Mean recent raw cross-context discrepancy & Recent contradiction statistic & Used by the Crucible reversal test \\[3pt]

$\Dscr_{\mathrm{var}}$ & Rolling discrepancy variance & Variance over recent local history & Drives responsiveness\\[3pt]

$n_i$ & Exposure count & Number of local update events & Crystallization eligibility \\[3pt]

$n_{\mathrm{cross}}$ & Cross-context exposure count & Number of contradiction-test updates & Crucible eligibility \\[3pt]

$\theta_{\mathrm{conv}}$ & Convergence threshold & Recent-discrepancy threshold for crystallization & Stability criterion \\[3pt]

$\theta_{\mathrm{rev}}$ & Reversal threshold & Crucible dissolution threshold & False-stability criterion \\[3pt]

$\eta_i$ & Modification rate & Transient or crystallized write rate & Governs same-context spatial learning \\[3pt]

$w_{\mathrm{target}}$ & Responsiveness target & Variance-derived target value for $w_i$ & Low variance raises commitment; high variance lowers it \\[3pt]

\bottomrule
\end{tabular*}
\end{table*}

\end{document}